\documentclass[runningheads]{llncs}

\usepackage{eccv}

\usepackage{eccvabbrv}

\usepackage{graphicx}
\usepackage{booktabs}
\usepackage{subcaption}
\usepackage{enumitem}

\usepackage[accsupp]{axessibility}  

\usepackage[pagebackref,breaklinks,colorlinks,citecolor=eccvblue]{hyperref}

\usepackage{orcidlink}
\usepackage{pifont}
\usepackage{multirow}
\usepackage{booktabs}
\usepackage{multirow}
\usepackage{graphicx}
\usepackage{tikz}
\usetikzlibrary{calc}

\newcommand{\imgsrc}[4][\fontsize{2.6}{2.9}\selectfont]{%
\begin{tikzpicture}
\node[inner sep=0] (img) {\includegraphics[width=#2]{#3}};
\node[
    anchor=south east,
    text=black,
    opacity=0.55,
    font=#1,
    inner sep=0.35pt,
    align=right
] at ([xshift=-0.3pt,yshift=1.1pt]img.south east) {#4};
\node[
    anchor=south east,
    text=white,
    font=#1,
    inner sep=0.35pt,
    align=right
] at ([xshift=-0.7pt,yshift=0.7pt]img.south east) {#4};
\end{tikzpicture}%
}

\newcommand{\qualimgsrc}[5][\fontsize{2.35}{2.55}\selectfont]{%
\begin{tikzpicture}
\node[inner sep=0] (img) {\includegraphics[width=#2]{#3}};

\node[
    anchor=south east,
    text=black,
    opacity=0.55,
    font=#1,
    inner sep=0.25pt,
    align=right
] at ([xshift=-3.0pt,yshift=2.0pt]$(img.south west)!0.62!(img.south east)$) {#4};
\node[
    anchor=south east,
    text=white,
    font=#1,
    inner sep=0.25pt,
    align=right
] at ([xshift=-3.4pt,yshift=1.6pt]$(img.south west)!0.62!(img.south east)$) {#4};

\node[
    anchor=south east,
    text=black,
    opacity=0.55,
    font=#1,
    inner sep=0.25pt,
    align=right
] at ([xshift=-2.0pt,yshift=2.0pt]img.south east) {#5};
\node[
    anchor=south east,
    text=white,
    font=#1,
    inner sep=0.25pt,
    align=right
] at ([xshift=-2.4pt,yshift=1.6pt]img.south east) {#5};
\end{tikzpicture}%
}

\begin{document}

\title{OpenCVL: An Open, Diverse, and Large-Scale Dataset for Fine-Grained Cross-View Localization} 

\titlerunning{OpenCVL}

\author{Zimin Xia\inst{1}\inst{2}\textsuperscript{*}$^\dagger$\orcidlink{0000-0002-4981-9514} \and
Mubariz Zaffar\inst{3}\textsuperscript{*}\orcidlink{0000-0002-9368-2391} \and
Junsheng Fu\inst{4}\orcidlink{0000-0002-8111-256X} \and
Alexandre Alahi\inst{1}\orcidlink{0000-0002-5004-1498} \and
Julian F. P. Kooij\inst{3}\orcidlink{0000-0001-9919-0710}}

\authorrunning{Z.~Xia et al.}

\institute{École Polytechnique Fédérale de Lausanne (EPFL), Switzerland 
\and
Southern University of Science and Technology
 (SUSTech), China 
\and
Delft University of Technology, The Netherlands
 \and
Zenseact\\
\textsuperscript{*}Equal contribution, $^\dagger$corresponding author \\
Project homepage: \url{https://open-cvl.github.io/}
}

\maketitle


\begin{abstract}
  Fine-grained Cross-View Localization (CVL) estimates the precise position and orientation of a ground-level image by aligning it with geo-referenced aerial imagery, offering a scalable alternative to Global Navigation Satellite Systems (GNSS) in challenging urban environments. Existing datasets rely on data collected with high-end sensor suites, which inherently limit image diversity and scalability. While in-the-wild images are abundant, their noisy geo-tags make them unsuitable for reliable evaluation. To bridge this gap, we introduce OpenCVL, a large-scale, diverse, and open dataset containing 617,388 ground-aerial image pairs spanning 41 cities across four European countries. All images are sourced from permissive platforms, ensuring long-term accessibility and supporting open and reproducible research. The training set combines images captured with high-end sensors with diverse in-the-wild imagery. We further develop a data curation framework that filters and corrects pose annotations to construct reliable in-the-wild evaluation data. In addition, OpenCVL includes dedicated cross-area and snowy test sets to assess generalization and robustness. Experiments with a state-of-the-art CVL model on OpenCVL show that incorporating noisy in-the-wild data consistently improves performance on clean test sets, suggesting a promising direction for scaling CVL with diverse real-world imagery.
  
\end{abstract}

\section{Introduction}
\label{sec:intro}

\begin{figure}[t]
    \centering
    \includegraphics[width=1.0\linewidth]{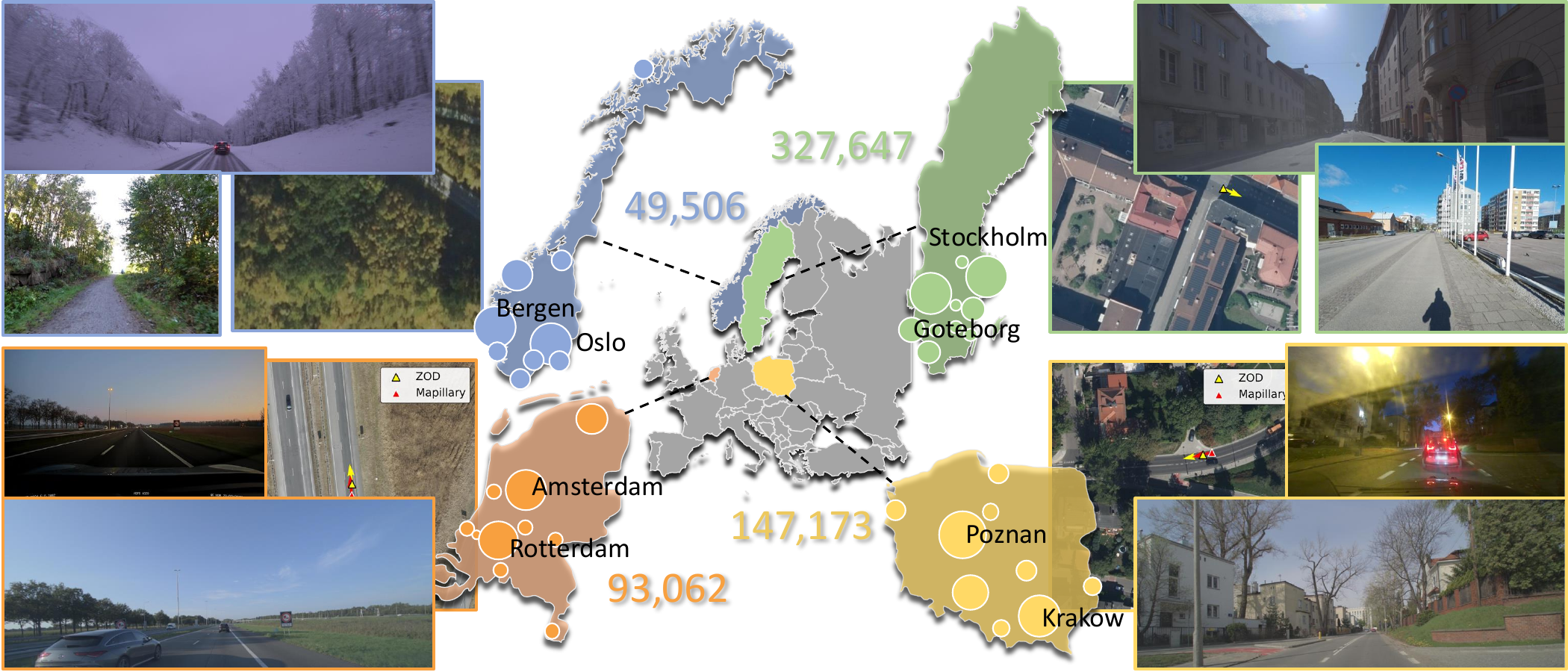}
    \caption{Overview of \textbf{OpenCVL}, an \textbf{Open} dataset for fine-grained \textbf{Cross-View Localization}, spanning 41 cities across four {Euro}pean countries and covering more than 7,000 km$^2$. The dataset encompasses diverse geographic regions and provides substantial variability in camera types, viewpoints, as well as temporal and weather conditions.
    The ground-level images are collected from Mapillary~\cite{Mapillary}
    (CC BY-SA) and ZOD~\cite{alibeigi2023zenseact} (CC BY-SA).
    The aerial imagery are sourced from open data from Lantmäteriet~\cite{Sweden} (Sweden, CC BY 4.0), PDOK~\cite{Netherlands}
    (Netherlands, CC BY 4.0), Kartverket/Geovekst~\cite{Norway}
    (Norway, Åpne data, Norge digitalt-lisens), and GUGiK Geoportal~\cite{Poland}
    (Poland, INSPIRE noConditionsApply / noLimitations).
    }
    \label{fig:intro}
\end{figure}

Fine-grained Cross-View Localization (CVL) aims to estimate the precise geographic location and orientation of a ground-level query image by aligning it with geo-referenced aerial or satellite imagery of its surroundings~\cite{shi2022beyond,xia2022visual,xia2023convolutional,fervers2023uncertainty,wang2024fine,song2024learning,xia2025fg}.
By enabling meter-level positioning without reliance on detailed 3D city models, this paradigm provides a scalable alternative or complement to GNSS, particularly in environments where GNSS suffers from multipath effects~\cite{zhu2018gnss}, such as urban canyons and dense city centers, making it highly promising for self-driving, mobile robotics, and visual navigation.

The rapid advancement of fine-grained CVL has been largely enabled by the availability of benchmark datasets~\cite{zhu2021vigor,shi2022beyond,xia2022visual}.
In these datasets, accurate ground-truth poses are obtained using dedicated data-collection vehicles equipped with high-end sensor suites~\cite{Geiger2013IJRR,maddern20171,agarwal2020ford}.
While such carefully curated data has driven steady algorithmic progress, this acquisition paradigm inherently constrains data diversity in terms of geographic coverage, collection time, as well as camera types and mounting configurations, and is therefore suboptimal for scaling fine-grained CVL to diverse real-world use cases.

On the other hand, in-the-wild images, captured with diverse cameras under unconstrained time, locations, and viewing directions, are abundant, e.g., from the Mapillary~\cite{Mapillary} platform, but their associated geo-tags are noisy due to reliance on phone-grade GNSS.
Prior work~\cite{fervers2024statewide} has leveraged such noisy data to advance large-scale coarse localization, where a prediction is deemed correct if it lies within a $50~\text{m}$ radius of the true position. 
However, such coarse ground-truth pose labels are inadequate for evaluating fine-grained CVL. 
Moreover, it remains unclear whether noisy crowd-sourced data can benefit the training of fine-grained CVL models when combined with less diverse but more accurate data.

Recently, non-public proprietary data~\cite{lindenberger2025scaling,sarlin2024snap} has been used to scale up CVL. However, access-restricted download protocols prevent independent validation and often impose limitations on downloading, sharing, and modifying such datasets, which can hinder reproducibility and the development of new research ideas.
In fact, existing benchmark datasets~\cite{zhu2021vigor,shi2022beyond,xia2022visual} contain images sourced from Google Maps or Google Street View, which are distributed under restrictive terms of use. While these datasets are widely used in current research, the licensing restrictions pose challenges for long-term accessibility and fully open, reproducible research.

To address this gap, we introduce {OpenCVL}, a fine-grained CVL dataset spanning four {European countries}. We meticulously ensure three characteristics for our dataset: open, diverse, and large-scale. OpenCVL provides high-resolution aerial imagery with ground-level images captured under heterogeneous sensing conditions (see Fig.~\ref{fig:intro}).
All ground-level and aerial images in the {OpenCVL} dataset are sourced from permissive platforms, including the Zenseact Open Dataset (ZOD)~\cite{alibeigi2023zenseact}, Mapillary~\cite{Mapillary}, and national mapping agencies that provide open access to aerial imagery.
The training set contains two complementary subsets: one collected with vehicle-mounted high-end sensors to provide accurate pose supervision, and another comprising in-the-wild data with noisier poses but substantially greater image diversity. 
Importantly, for both subsets, highly accurate test splits are created to ensure reliable evaluation. 
A dedicated framework has been developed that automatically filters and corrects the poses of in-the-wild Mapillary imagery.
Through these test splits, we define multiple challenges to assess cross-area generalization, robustness to seasonal variations, and performance under in-the-wild conditions.

Concretely, our main contributions are:
(i) We propose a large-scale and diverse dataset for fine-grained CVL. It is the first dataset in this field where all data originates from permissive sources, ensuring long-term accessibility and supporting open research.
(ii) We develop a data curation framework that automatically filters and corrects the ground-truth poses of in-the-wild ground-level images. This framework uniquely enables the construction of a challenging, highly accurate in-the-wild test set, addressing the lack of in-the-wild evaluation data in existing CVL datasets. Benchmarking state-of-the-art CVL methods on this in-the-wild test set reports a large room for improvement in existing research.
(iii) We provide baseline results on our dataset to facilitate future comparisons. Our experiments show that training with noisy in-the-wild data improves localization accuracy on all test sets, suggesting a promising direction for scaling CVL with noisy but diverse real-world data.

\section{Related Work}
\label{sec:relatedwork}

\textbf{Cross-view localization} aims to determine the geographic location of a query ground-level image by matching it to georeferenced aerial imagery. This task is commonly divided into image retrieval~\cite{hu2018cvm,ground-to-aerialgeolocalization} and fine-grained localization~\cite{xia2022visual}.

Cross-view image retrieval targets large-scale, coarse localization. 
It matches a ground-level query image against a pre-constructed aerial database covering a known geographic region, e.g., a city~\cite{vo2016localizing,liu2019lending,fervers2024statewide} or a country~\cite{WideAreaImageGeolocalization,zhai2017predicting}, with the objective of identifying the aerial image that covers the query location. 
Fine-grained CVL narrows the problem scope and aims to estimate the precise location and orientation of the query within an aerial image. 
Some approaches match the ground view to patch-level aerial descriptors~\cite{lentsch2023slicematch,xia2022visual,xia2023convolutional}, while others project the ground view into bird’s-eye-view space to facilitate geometric alignment with aerial imagery~\cite{fervers2023uncertainty,wang2023view,wang2024fine,wang2025bevsplat,wang2024view,xia2025fg}. 
Recently, Loc$^2$~\cite{xia2025loc} demonstrated that dense local feature correspondences between ground and aerial images can also be directly established. In addition to those fully supervised deep models, several works explore weakly supervised learning strategies to reduce reliance on precise ground-truth annotations~\cite{xia2024adapting,shi2024weakly,wang2025bevsplat,tong2025geodistill}.

\textbf{Cross-view datasets} have played a crucial role in advancing CVL by pairing ground-level images with georeferenced aerial or satellite imagery. 
Most datasets are primarily designed for cross-view image retrieval~\cite{lin2013cross,WideAreaImageGeolocalization,vo2016localizing,tian2017cross,zhai2017predicting,liu2019lending,fervers2024statewide}. 
While these datasets offer ground–aerial image pairs, they typically lack precise ground-truth pose annotations for the ground cameras.
VIGOR~\cite{zhu2021vigor} is the first dataset to support fine-grained CVL. 
It contains geo-tagged Google Street View panoramas paired with Google Maps satellite images, covering four cities in the US.
Although panoramic images provide richer scene context that can ease cross-view image matching, most consumer devices capture perspective images (e.g., from mobile phone cameras).
Subsequent datasets~\cite{shi2022beyond,xia2022visual} are often constructed on top of existing driving datasets, such as KITTI~\cite{Geiger2013IJRR}, FordAV~\cite{agarwal2020ford}, and Oxford RobotCar~\cite{maddern20171}. However, these datasets remain limited in camera types, viewing directions, and geographic diversity, and their aerial imagery is also sourced from Google Maps, which is distributed under restrictive terms of use.
SNAP~\cite{sarlin2024snap} also targets large-scale CVL, but it relies on proprietary data. 
Currently, there is a lack of an open, large-scale, and diverse dataset in this field.
\section{The OpenCVL Dataset}
\label{sec:dataset}

We propose OpenCVL, the largest and most diverse open dataset to date
for training and evaluating fine-grained CVL models.
Its goals are: 
(i) scale model training with large and diverse data, 
(ii)~provide accurate pose labels for diverse (even in-the-wild) ground-level imagery for evaluation.
For (ii), we develop a framework to automatically filter and correct labels of in-the-wild imagery.

\subsection{Data Sources}
OpenCVL's ground-level images are collected from two complementary sources, ZOD and Mapillary, both released under permissive CC BY-SA licenses.
Four countries were selected that are covered by both ZOD and Mapillary and whose national mapping agencies provide geo-referenced high-resolution aerial imagery with public open access: Sweden, Poland, Norway, and the Netherlands.

\textbf{Zenseact Open Dataset (ZOD):}
ZOD~\cite{alibeigi2023zenseact} was collected using vehicle-mounted sensor rigs equipped with a front-facing camera, a LiDAR, and a high-end GNSS across multiple European countries. The dataset includes standalone frames and video sequences in two formats, i.e., 20-second clips and several-minute drives. We adopt both the frames and videos in our selected countries.

\textbf{Mapillary:}
Mapillary~\cite{Mapillary} is an open platform hosting crowd-sourced street-level imagery from across the world. The images are captured using a wide range of camera devices, different platforms (cars, pedestrians, cyclists, etc.), and across diverse times and locations. Due to its crowd-sourced nature, this imagery provides diversity unseen in other driving datasets, though the GNSS position estimates are noisy. Each image is associated with a phone-grade raw GNSS tag and a heading, a so-called improved GNSS tag and heading computed via OpenSfM~\cite{adorjan2016opensfm}, and other metadata including the camera intrinsics and distortion parameters. 
However, both raw GNSS tags and OpenSfM poses remain noisy and are insufficient for benchmarking CVL (see details in Sec.~\ref{sec:impactofmapillarydatacuration}).\\

\textbf{Aerial Imagery:}
For each ground-level image, we retrieve a corresponding aerial image covering a $100~\text{m} \times 100~\text{m}$ area at the highest available resolution from the national mapping agencies of Sweden~\cite{Sweden}, Poland~\cite{Poland}, Norway~\cite{Norway}, and the Netherlands~\cite{Netherlands}.
Similar to other fine-grained CVL datasets~\cite{zhu2021vigor,shi2022beyond,xia2022visual}, the ground-view camera location is \textit{not} placed at the center of the aerial image.
Instead, it is randomly located within a $40\,\mathrm{m}\times40\,\mathrm{m}$ region centered in the aerial image, corresponding to offsets of up to $\pm 20$~m along both the east--west and north--south map axes.
All aerial images are north-aligned and have a ground sampling distance ranging from 0.04~m/pixel to 0.16~m/pixel (see Tab.~\ref{tab:dataset_statistics}).

\begin{table}[t]
    \centering
    \caption{Statistics of the proposed OpenCVL dataset across target countries.}
    \label{tab:dataset_statistics}
    \begin{tabular}{lcccc}
        \toprule
        & Sweden & Poland & Norway & Netherlands \\
        \midrule
        Ground-level: ZOD (\# images) & 241107 & 29663 & 1403 & 2314 \\
        Ground-level: Mapillary (\# images) & 86540 & 117510 & 48103 & 90748 \\
        Aerial Image Resolution (m/pixel) & 0.16 & 0.1 & 0.04--0.1 & 0.045 \\
        \bottomrule
    \end{tabular}
\end{table}

\subsection{Data Splits}
The OpenCVL dataset is divided into training, validation, and test sets. 
For the training set, we aim to ensure both reliable supervision and high data diversity. 
To this end, we combine the more accurate ZOD data with diverse but noisier in-the-wild imagery from Mapillary, covering a wide range of camera types, viewpoints, and capture conditions. Although the Mapillary poses are noisy, our experiments show that incorporating this data significantly benefits training due to its broader image distribution. 
Furthermore, this training setup is more scalable than the conventional paradigm that relies solely on carefully curated datasets but sacrifices data scale and diversity.
 
For evaluation, accurate pose annotations are required. 
Therefore, the validation and test sets are carefully curated to ensure reliable ground-truth poses. 
We further introduce several evaluation challenges to assess model robustness under different conditions: cross-area generalization to unseen locations, a snowy test set for seasonal appearance changes, and an in-the-wild test set reflecting diverse viewpoints and sensing conditions. 

In total, the dataset contains 617,388 ground-aerial image pairs, of which 579,752 are used for training, including 238,212 from ZOD and 341,540 from Mapillary. The validation, cross-area, snowy, and in-the-wild test sets contain 14,756, 18,504, 3,015, and 1,361 image pairs, respectively.
Next, we describe how these validation and test splits are constructed.

\subsection{Creating Accurate Validation and Test Splits} 
\label{sec:test_splits}

\subsubsection{From ZOD:} 
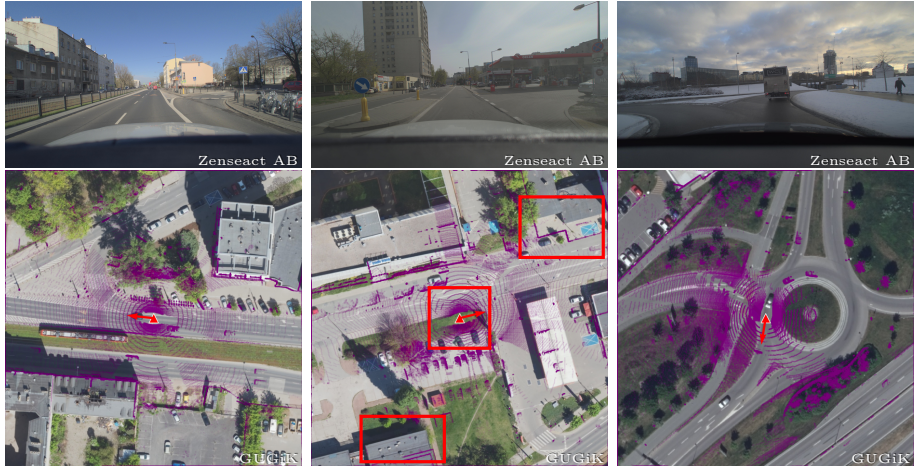
\begin{figure}[t]
    \centering
    \setlength{\tabcolsep}{2pt}
    \renewcommand{\arraystretch}{0}

    \begin{tabular}{ccc}
        \imgsrc[\fontsize{3.2}{3.5}\selectfont]{0.32\linewidth}
        {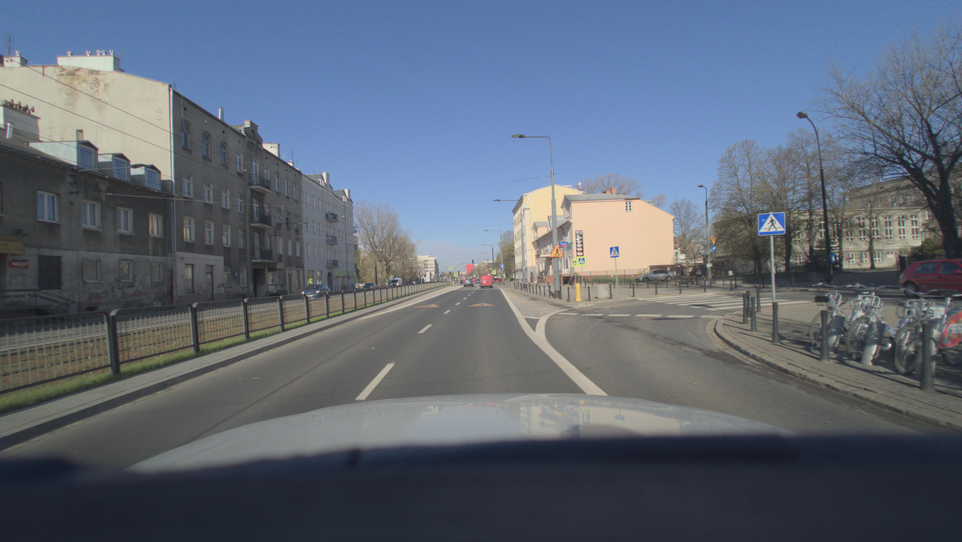}
        {Zenseact AB} &
        \imgsrc[\fontsize{3.2}{3.5}\selectfont]{0.32\linewidth}
        {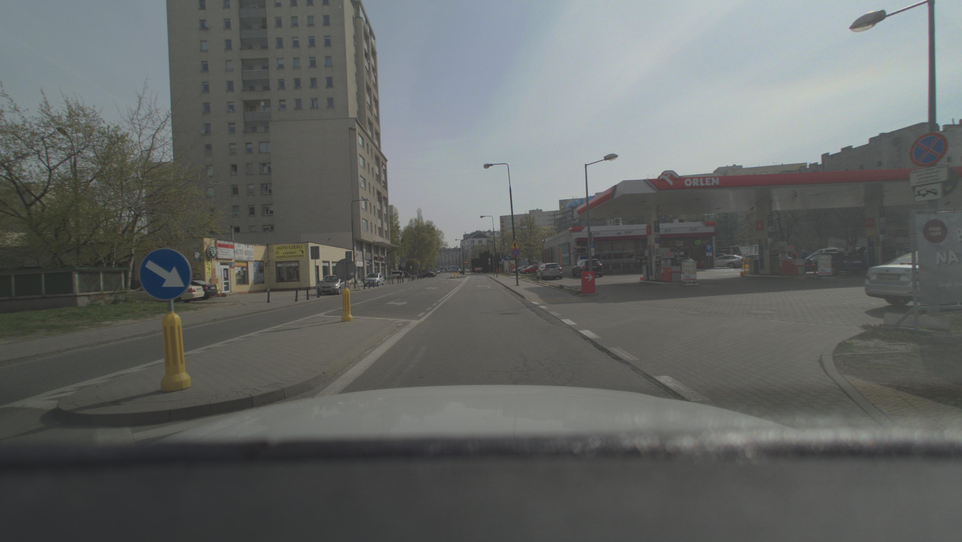}
        {Zenseact AB} &
        \imgsrc[\fontsize{3.2}{3.5}\selectfont]{0.32\linewidth}
        {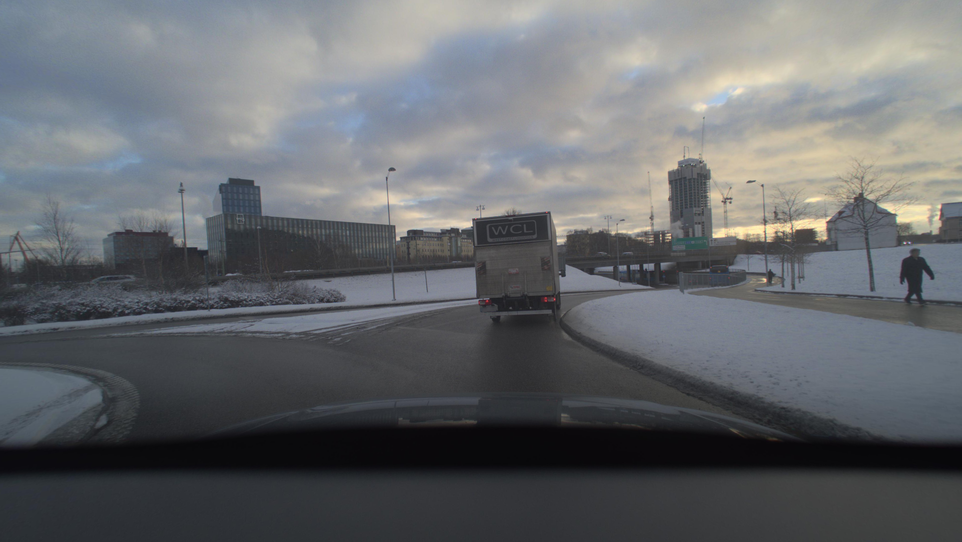}
        {Zenseact AB}
        \\[1pt]

        \imgsrc[\fontsize{3.2}{3.5}\selectfont]{0.32\linewidth}
        {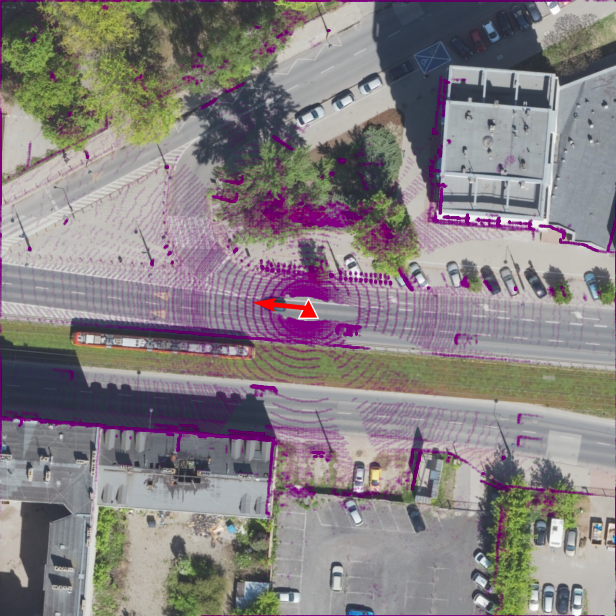}
        {GUGiK} &
        \begin{tikzpicture}
        \node[inner sep=0] (img) {\includegraphics[width=0.32\linewidth]{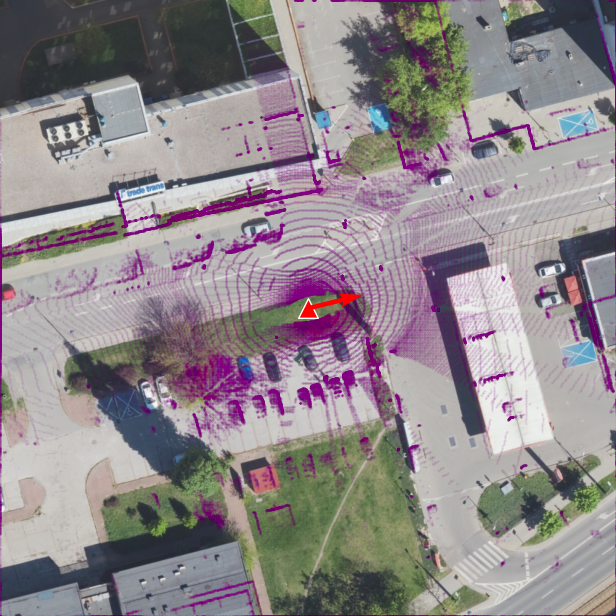}};
        \node[anchor=south east, text=black, opacity=0.55,
              font=\fontsize{3.2}{3.5}\selectfont, inner sep=0.35pt, align=right]
              at ([xshift=-0.3pt,yshift=1.1pt]img.south east) {GUGiK};
        \node[anchor=south east, text=white,
              font=\fontsize{3.2}{3.5}\selectfont, inner sep=0.35pt, align=right]
              at ([xshift=-0.7pt,yshift=0.7pt]img.south east) {GUGiK};
        \draw[red, line width=1.2pt] (-0.4,-0.4) rectangle (0.4,0.4);
        \draw[red, line width=1.2pt] (0.8,0.8) rectangle (1.9,1.6);
        \draw[red, line width=1.2pt] (-0.2,-1.3) rectangle (-1.3,-1.9);
        \end{tikzpicture} &
        \imgsrc[\fontsize{3.2}{3.5}\selectfont]{0.32\linewidth}
        {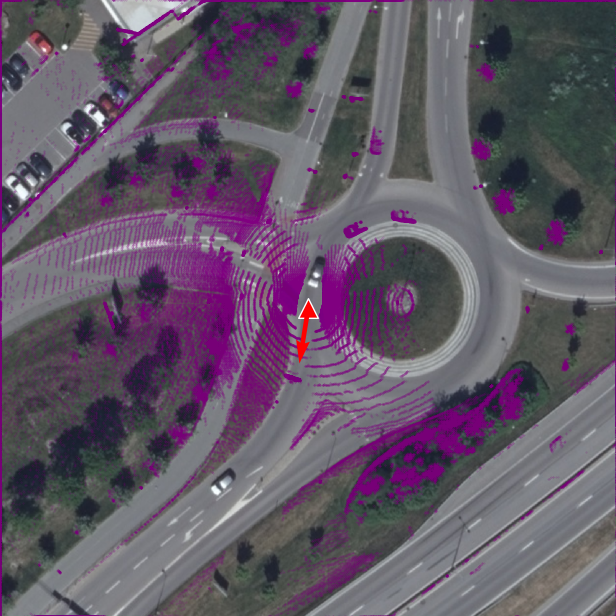}
        {GUGiK}
    \end{tabular}

    \caption{Examples of pose verification in OpenCVL test sets curated from ZOD. We verify the projected camera poses and LiDAR point cloud (shown in purple) on the aerial imagery. \textbf{Left:} good alignment. \textbf{Middle:} misalignment, where the red boxes highlight the incorrect camera location and the mismatch between building edges and LiDAR points. \textbf{Right:} examples of selected snowy samples with good alignment.}
    \label{fig:zod_lidar}
\end{figure}

To construct clean validation and test sets from ZOD, we manually verify the accuracy of the camera poses for the selected evaluation samples by projecting them together with the recorded LiDAR point clouds into aerial imagery and inspecting their geometric alignment, see Fig.~\ref{fig:zod_lidar}. Sequences with noticeable pose errors are filtered out.

We create two test splits from ZOD: a \emph{cross-area} test set, and a \emph{snowy} test set.
The \emph{cross-area} test set is created by selecting sequences from geographic regions that do not overlap with the training data. The \emph{snowy} test set is obtained by selecting scenes recorded under winter conditions, where snow is visible on the ground or the surrounding environment.
Note that the training data also includes winter scenes. As such data can be easily collected in practice, our objective is not cross-season generalization,
but to test robustness to such variations.

\subsubsection{From Mapillary:} 
We aim to construct a clean test set from crowd-sourced Mapillary imagery with accurate ground-truth poses for evaluating CVL methods. The raw GNSS tags and the OpenSfM-refined poses provided by Mapillary are often inaccurate and therefore unsuitable for reliable evaluation, motivating the need for a dedicated pose-correction framework.
Importantly, ZOD also provides LiDAR point clouds, which in principle offer sufficient geometric information to estimate accurate poses for nearby Mapillary images by registering them to the globally aligned 3D point clouds.

Therefore, we retrieve Mapillary images around ZOD coverage areas based on their raw GNSS tags. Visual overlap between the ZOD and Mapillary images is then verified using local feature matching with geometric verification. Images with insufficient geometrically verified matches are automatically discarded.

\begin{figure}[t]
    \centering
    \includegraphics[width=1.0\linewidth]{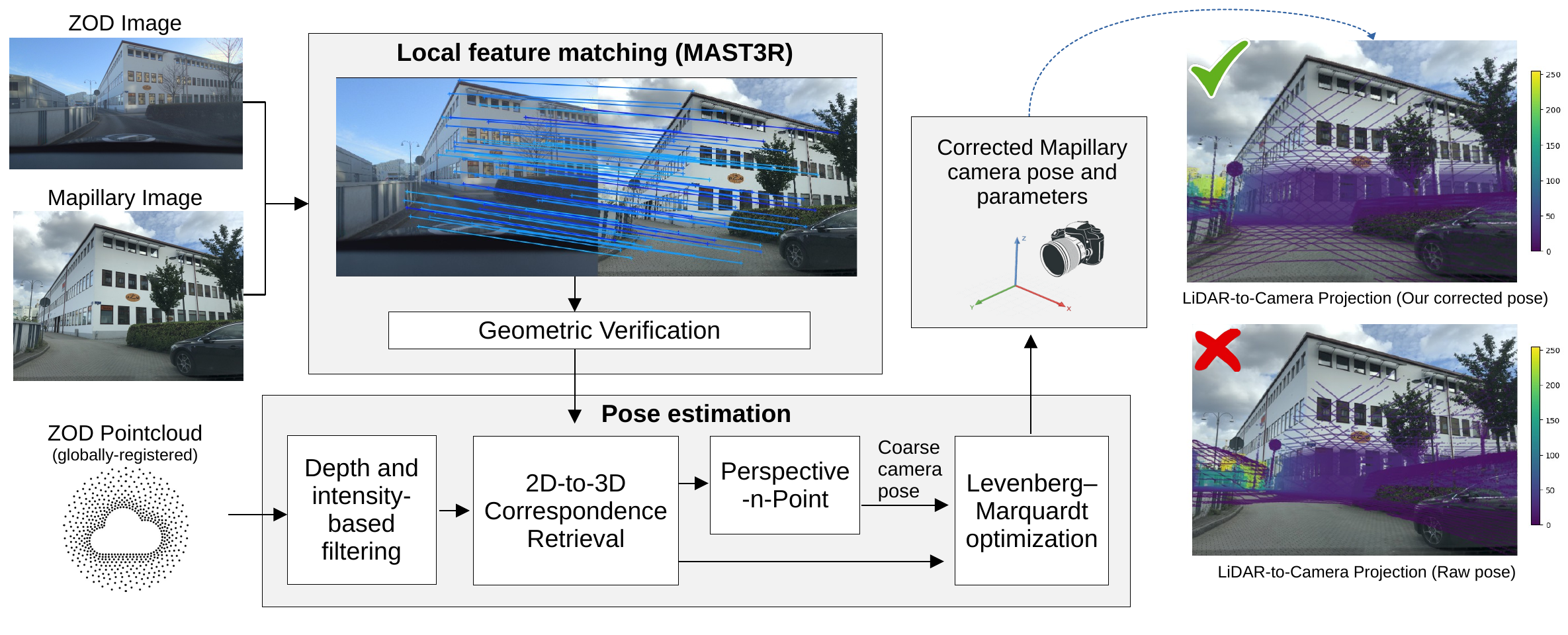}
    \caption{Our framework for estimating accurate poses for Mapillary images. Feature matching is performed between an anchor ZOD image and the Mapillary image, followed by pose estimation using 2D--3D correspondences obtained from the globally registered ZOD LiDAR point cloud. The quality of the estimated pose is illustrated on the right by projecting the ZOD point cloud into the Mapillary image using our refined pose compared to the raw Mapillary GNSS pose.}
    \label{fig:eccv_2026_pipeline}
\vspace{-3mm}

\end{figure}

For images with good visual overlap, the accurate pose and camera intrinsics for each Mapillary image are then computed using our framework described in Fig.~\ref{fig:eccv_2026_pipeline}. Concretely, we compute local feature matches between the ZOD and Mapillary image using the state-of-the-art local feature extractor MAST3R~\cite{leroy2024grounding} with reciprocal matching and geometric verification to filter outliers. 
Since each ZOD image is accompanied by a corresponding LiDAR point cloud, we use the poses provided by the high-end GNSS system to register these point clouds globally by transforming each 3D point into global coordinates.
Given the local feature matches between ZOD and Mapillary images, we thus get a number of 2D-to-3D correspondences for the pixels in the Mapillary image to 3D points in global coordinates. 
These correspondences are then used first to estimate a coarse Mapillary camera pose using Perspective-n-Point (PnP), which is further refined along with the camera intrinsics and distortion parameters using Levenberg–Marquardt optimization. We use the COLMAP~\cite{schoenberger2016sfm} framework for this pose estimation. The quality of pose estimation is then determined via the reprojection error, such that the pose estimates with higher reprojection error are automatically discarded.

To further ensure the high-quality of the ground-truth pose labels, the computed Mapillary poses were verified manually: a) by visualizing the LiDAR projection into the Mapillary image using the computed pose (see Fig.~\ref{fig:eccv_2026_pipeline} right), and b) by verifying the computed pose in the aerial image (see Fig.~\ref{fig:examples} bottom). Fig.~\ref{fig:eccv_2026_pipeline} also shows the quality of the ground-truth pose computed by our framework in comparison to the reported Mapillary raw GNSS pose.




\subsection{Comparison to Prior Datasets}
We here present a direct comparison between the proposed OpenCVL dataset and the existing fine-grained CVL datasets, see Tab.~\ref{tab:dataset_comparison}.

\begin{figure}[t]
    \centering
    \setlength{\tabcolsep}{1.5pt}
    \renewcommand{\arraystretch}{0}

    \begin{tabular}{ccccc}
        \imgsrc{0.19\linewidth}{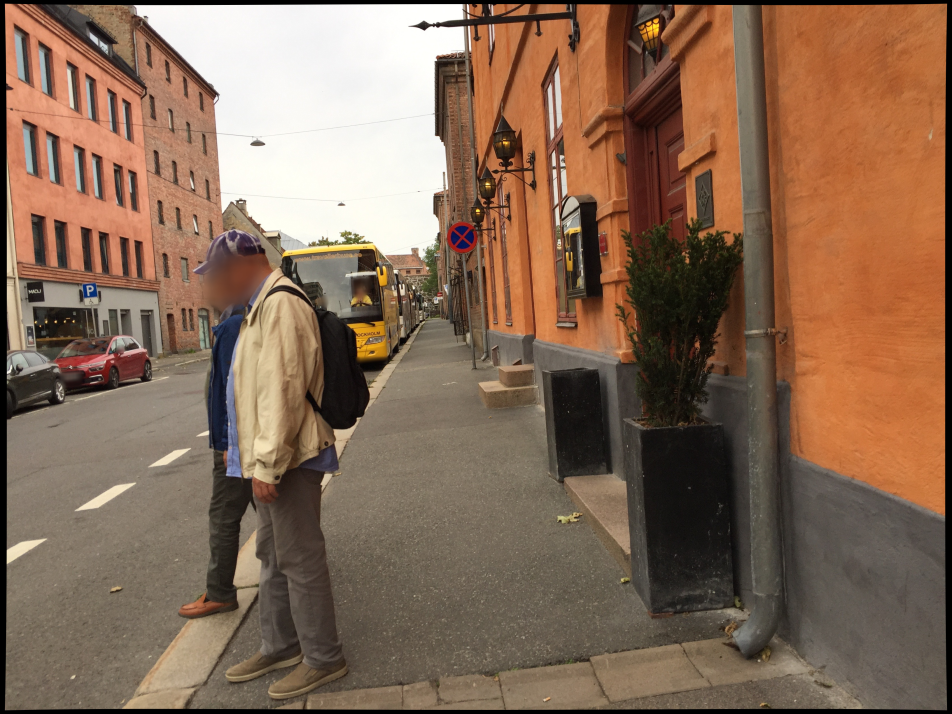}{Mapillary/richlv} &
        \imgsrc{0.19\linewidth}{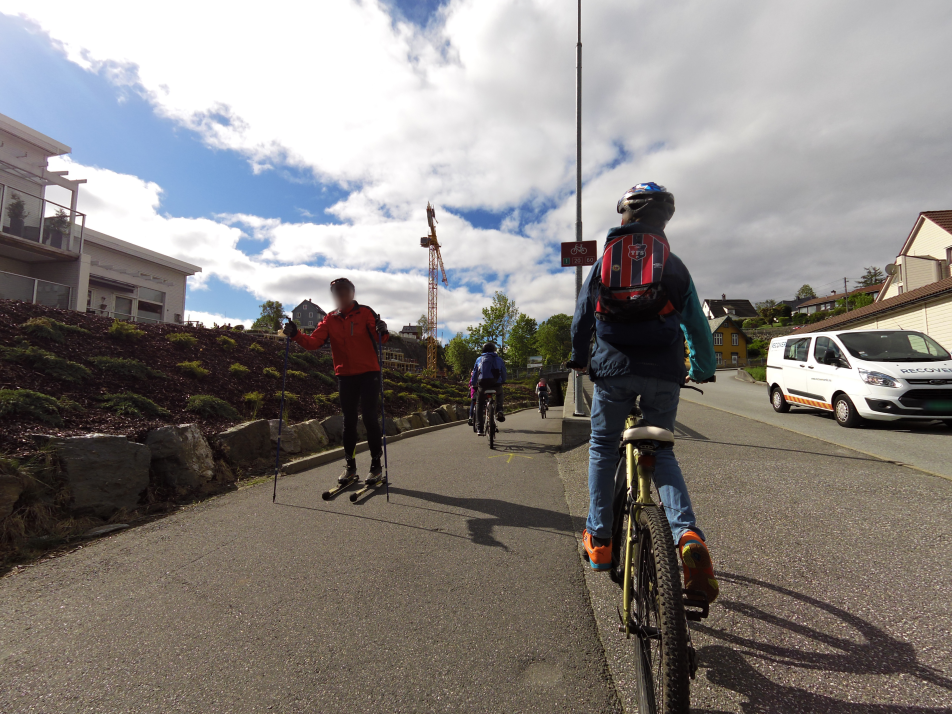}{Mapil./suschtenname} &
        \imgsrc{0.19\linewidth}{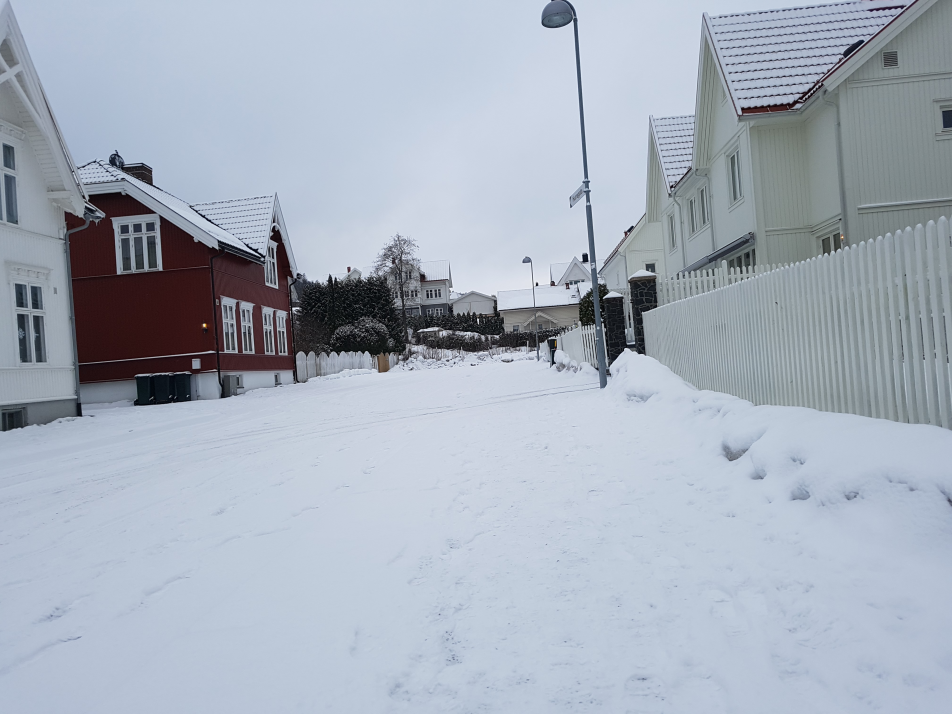}{Mapillary/pbb} &
        \imgsrc{0.19\linewidth}{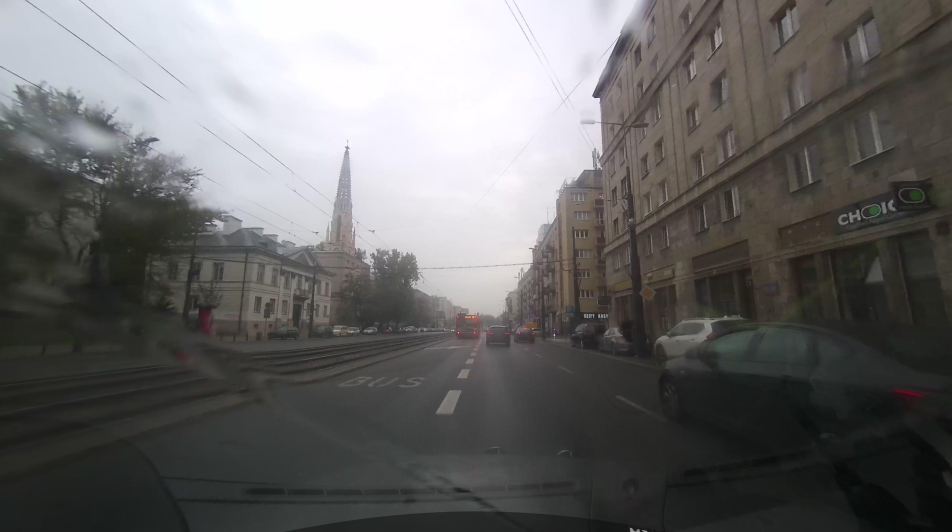}{Mapillary/grobin} &
        \imgsrc{0.19\linewidth}{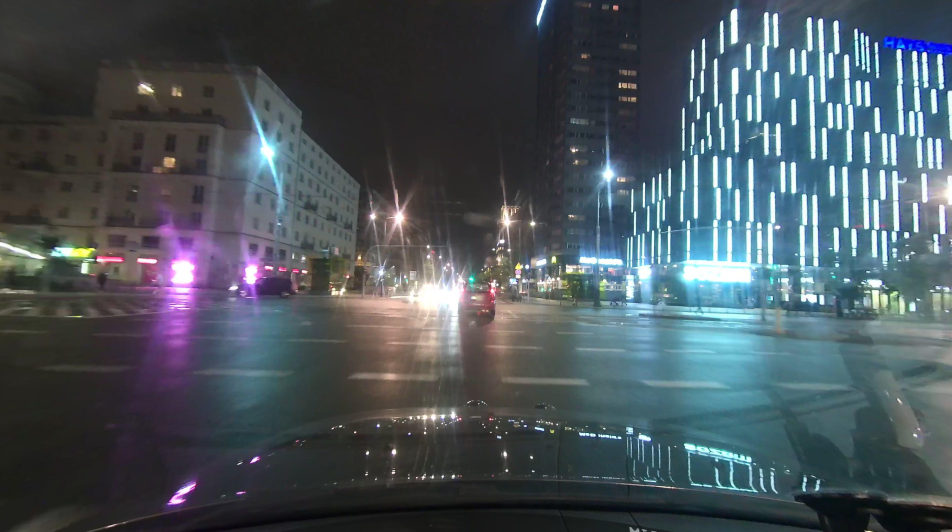}{Mapillary/grobin}
        \\[1pt]

        \imgsrc{0.19\linewidth}{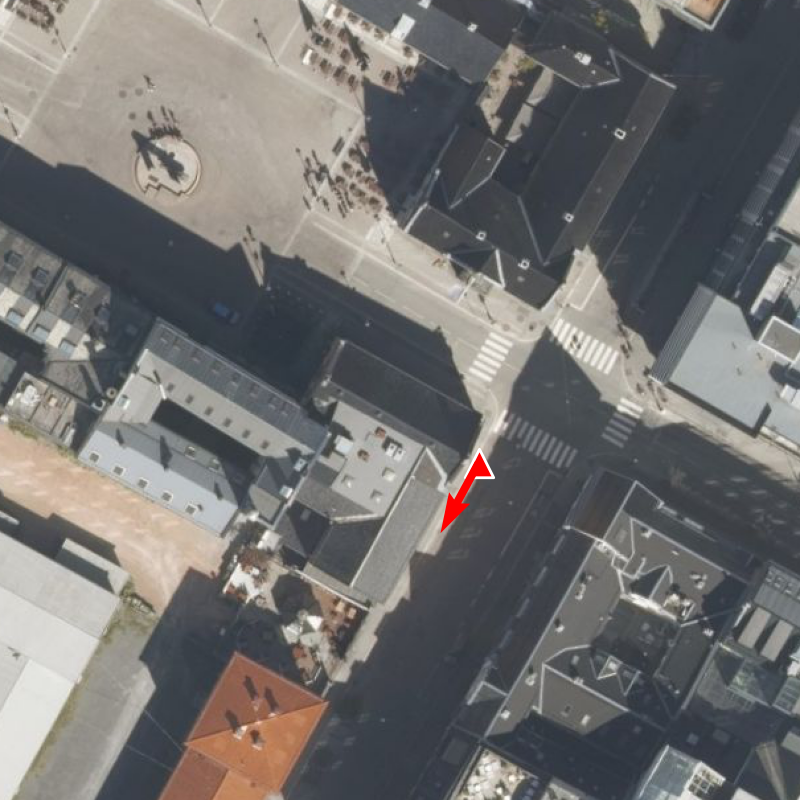}{Kartverket} &
        \imgsrc{0.19\linewidth}{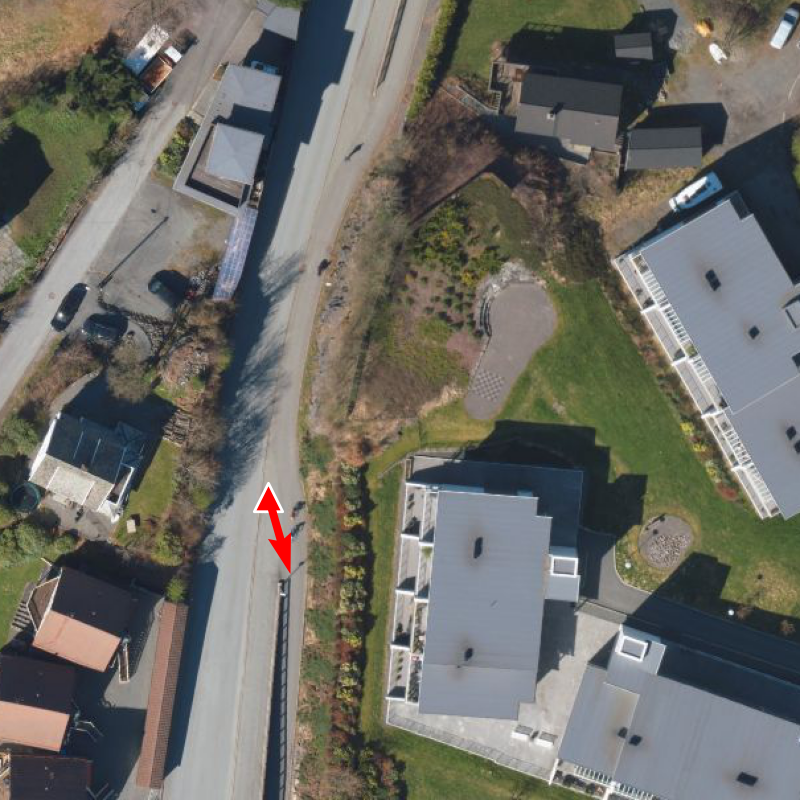}{Kartverket} &
        \imgsrc{0.19\linewidth}{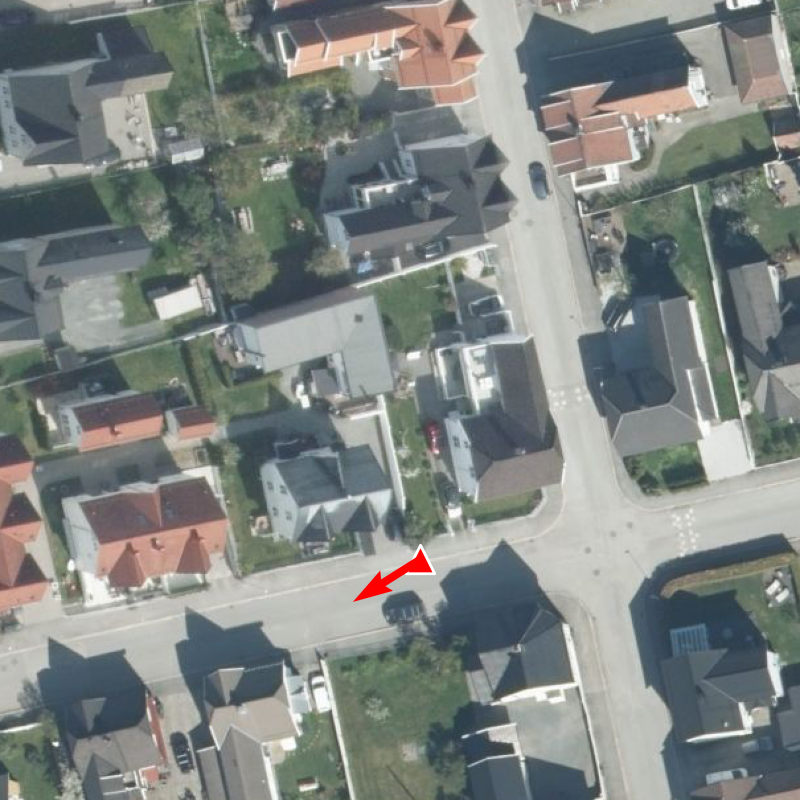}{Kartverket} &
        \imgsrc{0.19\linewidth}{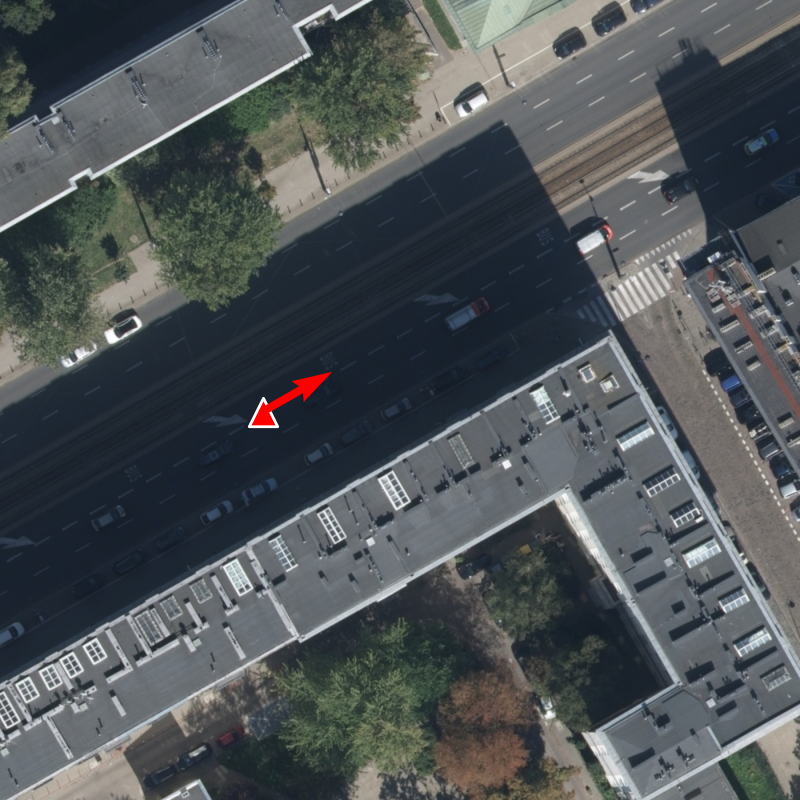}{GUGiK} &
        \imgsrc{0.19\linewidth}{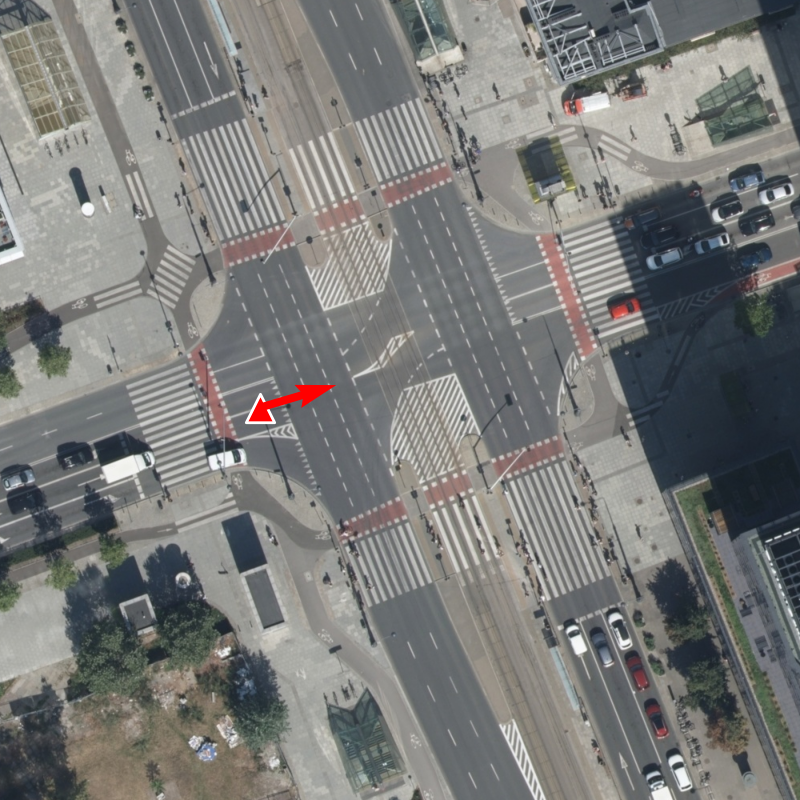}{GUGiK}
    \end{tabular}

    \caption{Examples of diverse ground images and their corresponding aerial views in the OpenCVL dataset. Our dataset uniquely contains ground-level images taken by pedestrians and cyclists in truly in-the-wild conditions.}
    \label{fig:examples}
\end{figure}

\textbf{Collection platform:}
In prior datasets, ground-level images are typically collected using vehicle-mounted platforms~\cite{Geiger2013IJRR,agarwal2020ford,maddern20171}. 
This setup restricts the data to on-road scenes with forward-facing viewpoints. Moreover, such datasets rely on fixed camera configurations, for example, a single camera type and a fixed mounting height, which does not reflect the variability encountered in real-world deployment scenarios.


In contrast, OpenCVL combines diverse camera types and mounting configurations.
While ZOD provides data captured using fixed, calibrated vehicle-mounted cameras,  Mapillary contributes images acquired from a wide range of devices, including handheld, bicycle-mounted, and dash cameras (see Fig.~\ref{fig:examples}). Due to these heterogeneous acquisition platforms, our dataset spans a broad range of viewpoints, including perspectives from sidewalks, cyclist lanes, and roadways.
Such diversity in camera types, mounting configurations, and viewpoints distinguishes is not found in prior fine-grained cross-view localization datasets.

\textbf{Temporal diversity:}
The ground-level images in VIGOR~\cite{zhu2021vigor} and KITTI~\cite{Geiger2013IJRR} are predominantly captured during daytime, typically under clear and sunny environments, exhibiting limited variation in lighting and weather conditions. 
OpenCVL instead includes imagery collected at different times of day, across diverse weather conditions and multiple seasons.
While RobotCar~\cite{maddern20171} and FordAV~\cite{agarwal2020ford} also capture temporal and seasonal variations, our dataset combines these with diversity in collection platforms and provides significantly wider geographic coverage.

\textbf{Geographic coverage:}
Most prior datasets are limited to relatively small areas, such as Karlsruhe (KITTI), Michigan (FordAV), or Oxford (RobotCar), which restricts their suitability for large-scale training. In contrast, our dataset covers 41 cities from four European countries: Sweden, Poland, Norway, and the Netherlands. 
Together, they provide substantial diversity in architectural styles, road layout, and environmental conditions, more faithfully reflecting the variability encountered in real-world fine-grained CVL scenarios.

\textbf{Image source:}
Another important distinction lies in the data source of the provided imagery. Although this aspect is less often discussed, the use of open data is important to guarantee  long-term accessibility of the data, as well as the reproducibility of results.
Prior datasets~\cite{shi2022beyond,xia2022visual} rely on imagery from Google Maps, and VIGOR~\cite{zhu2021vigor} additionally uses Google Street View. 
In contrast, all imagery in our dataset originates from sources released under permissive licenses, such as CC BY-SA, or provides public free use. As a result, OpenCVL can be freely shared and extended, supporting reproducible research.

\begin{table}[t]
\centering
\caption{Comparison of existing fine-grained CVL datasets to our OpenCVL dataset.}
\label{tab:dataset_comparison}
\scriptsize
\begin{tabular}{lccccc}
\toprule
Dataset & Camera Type & Mounting & Temporal Diversity & Coverage & Data Source\\
\midrule
VIGOR~\cite{zhu2021vigor} & Fixed & Vehicle \& backpack & Limited & 4 U.S. cities & Restricted\\
FordAV~\cite{shi2022beyond} & Fixed & Vehicle & Diverse & Michigan, U.S. & Restricted\\
KITTI~\cite{shi2022beyond} & Fixed & Vehicle & Limited & Karlsruhe, DE & Restricted\\
RobotCar~\cite{xia2022visual} & Fixed & Vehicle & Diverse & Oxford, UK & Restricted\\
\midrule
\textbf{OpenCVL} & Mixed & Mixed & Diverse & 4 EU countries & Open-access\\
\bottomrule
\end{tabular}
\end{table}
\section{Experiments}
\label{sec:experiments}

We first introduce the experimental setup of our work.
Then we describe how we scale the training of a state-of-the-art fine-grained CVL method to accommodate the diversity of our OpenCVL dataset, followed by detailed experimental results demonstrating the advantages of our dataset. The benefits of our pose correction framework are also reported.

\textbf{Datasets:} In addition to OpenCVL, we also use the widely adopted KITTI dataset~\cite{shi2022beyond} in our experiments.
We do not use VIGOR, as it contains only panoramas, which are not directly comparable to the perspective images in OpenCVL.
The original KITTI dataset~\cite{Geiger2013IJRR} provides front-facing images collected by a single mapping vehicle in Germany, with no geographic overlap with OpenCVL.
We follow the protocol of~\cite{shi2022beyond}, in which the dataset is divided into same-area and cross-area splits, and each ground-level image lies within the central 40~m $\times$ 40~m region of its aerial image. Furthermore, consistent with recent works~\cite{xia2025loc}, we also report results under two settings: with an orientation prior (with noise within $\pm 10^\circ$) during both training and testing, and without this prior.

\textbf{Evaluation metrics:} The mean and median localization and orientation errors are used as evaluation metrics. Localization error is measured as the Euclidean distance in meters between the predicted and ground-truth planar positions, while orientation error measures the absolute angular difference between the predicted and ground-truth yaw angles.

\textbf{Baseline details:} 
We adopt the recent state-of-the-art fine-grained CVL method, Loc$^2$~\cite{xia2025loc}, as our main baseline to evaluate the effectiveness of the proposed dataset.
We additionally evaluate HC-Net~\cite{wang2024fine} and CCVPE~\cite{xia2023convolutional} as secondary baselines, with results reported in the Appendix.
Loc$^2$ first establishes local feature correspondences between ground and aerial images, and then estimates the 3-DoF camera pose of the ground image via scale-aware Procrustes alignment~\cite{umeyama1991least}. 
The method is trained end-to-end using only camera pose supervision, and its scale-aware Procrustes alignment requires camera intrinsics and the depth map for each ground-level image to lift the correspondences into the bird’s-eye-view space. We follow the original implementation and hyperparameter settings of Loc$^2$~\cite{xia2025loc}, including the number of sampled aerial and ground points, loss weights, and the maximum depth threshold. The only modification is the learning rate, which we set to $4 \times 10^{-4}$ to accommodate training on 4 H100 GPUs with a total batch size of 896.

\subsection{Scaling Training in Fine-Grained CVL }

\textbf{Training preparation:}
Since our data originates from diverse cameras, we resize all ground-level images to a fixed resolution and adjust their camera intrinsics accordingly.
Because camera intrinsics are explicitly used in the geometric lifting step of Loc$^2$, the local feature matching stage does not need to be aware of camera-specific parameters. This decoupling enables the model to learn a general ground-aerial image matching representation that generalizes across different cameras.
Following the original setup of Loc$^2$, we use DepthAnythingV2~\cite{yang2024depthv2} to predict metric depth for all ground-level images in OpenCVL. For images sourced from ZOD, we further refine the scale of the predicted depth maps using LiDAR point clouds from ZOD by aligning the model’s depth predictions with LiDAR-measured depth.

\textbf{Data weighting:}
Since our training data consists of high-confidence supervision from ZOD and noisier supervision from Mapillary, we reduce the influence of the noisy samples during optimization by downweighting their loss contribution, following noise-aware reweighting strategies~\cite{arazo2019unsupervised}.
For simplicity, we apply a fixed global downweighting factor to the loss of all Mapillary samples. We test downweighting factors of $0.1$, $0.3$, $0.5$, and $0.7$ for the losses of all Mapillary samples and find that $0.3$ yields slightly better results.

\subsection{Impact of Training Data Diversity}

\begin{table}[t]
\centering
\caption{Evaluation of Loc$^2$~\cite{xia2025loc} on the OpenCVL test sets under different training settings. 
We compare models trained on KITTI with and without the orientation prior, training on OpenCVL using only ZOD data, joint training with additional Mapillary data without loss downweighting, and the final setting that applies a global downweighting factor of 0.3 to Mapillary samples in OpenCVL. 
Best in bold.}
\label{tab:eva_proposed_data}
\scriptsize
\begin{tabular}{lcccccccccccc}
\toprule
& \multicolumn{4}{c}{Cross-area test set} 
& \multicolumn{4}{c}{Snowy test set} 
& \multicolumn{4}{c}{In-the-wild test set} \\

\cmidrule(lr){2-5}
\cmidrule(lr){6-9}
\cmidrule(lr){10-13}

Training data
& \multicolumn{2}{c}{Loc. (m)} & \multicolumn{2}{c}{Ori. ($^\circ$)}
& \multicolumn{2}{c}{Loc. (m)} & \multicolumn{2}{c}{Ori. ($^\circ$)}
& \multicolumn{2}{c}{Loc. (m)} & \multicolumn{2}{c}{Ori. ($^\circ$)} \\

\cmidrule(lr){2-3}
\cmidrule(lr){4-5}
\cmidrule(lr){6-7}
\cmidrule(lr){8-9}
\cmidrule(lr){10-11}
\cmidrule(lr){12-13}

& Mean & Med. & Mean & Med.
& Mean & Med. & Mean & Med.
& Mean & Med. & Mean & Med. \\

\midrule
KITTI w. ori. prior 
& 20.07 & 18.14 & 80.65 & 80.49
& 17.30 & 15.62 & 79.75 & 65.04
& 17.12 & 15.15 & 86.25 & 85.03 \\

KITTI w/o. ori. prior
& 15.57 & 13.60 & 79.75 & 72.93
& 15.92 & 14.48 & 83.30 & 76.41
& 15.98 & 14.07 & 70.72 & 57.15 \\

OpenCVL (ZOD only)
& 7.55 & 4.50 & 16.20 & 5.74
& 8.04 & 5.74 & 8.87 & \textbf{3.26}
& 8.61 & 6.85 & 35.59 & 12.45 \\

OpenCVL (w/o. weig.)
& 6.87 & 4.78 & 13.01 & 6.34
& 8.04 & 6.05 & 10.91 & 5.34
& 8.16 & 6.47 & \textbf{25.87} & 12.93 \\

\hline
OpenCVL
& \textbf{6.72} & \textbf{4.30} & \textbf{11.52} & \textbf{5.24}
& \textbf{7.80} & \textbf{5.48} & \textbf{5.09} & 3.52
& \textbf{7.90} & \textbf{6.19} & 26.07 & \textbf{12.01} \\
\bottomrule
\end{tabular}
\end{table}

\textbf{KITTI as training data:}
Although Loc$^2$ achieves around $1~\text{m}$ mean localization error on the original KITTI same-area test set~\cite{xia2025loc}, it exhibits high localization and orientation errors across all evaluation splits on OpenCVL (see Tab.~\ref{tab:eva_proposed_data}).
KITTI’s images lack diversity in terms of camera types, captured scenes, and viewing points. 
As a result, evaluating models solely on KITTI does not fully reflect their practical usefulness. 
Similarly, models trained on KITTI may struggle to generalize beyond the narrow training data distribution.

When an orientation prior is present in training, Loc$^2$~\cite{xia2025loc} may exploit this bias and learn to associate specific regions of the ground-view frustum with fixed regions in the aerial image. 
Consequently, the model trained with this prior performs worse than that trained without it when evaluated on OpenCVL, where the camera orientation is not assumed to be known.

\textbf{OpenCVL (ZOD only) as training data:}
As expected, the model trained on ZOD achieves significantly better performance than the model trained on KITTI in Tab.~\ref{tab:eva_proposed_data}. 
When evaluating the ZOD-trained model across different test splits, the performance on the cross-area and snowy sets is similar, with the snowy subset exhibiting slightly higher localization error but lower orientation error. 
Note that the training data already contains a range of weather conditions, as our objective is not to explicitly evaluate cross-weather generalization. 
The lower orientation error on the snowy subset may be attributed to its reduced diversity in viewing directions (mostly highway), which simplifies orientation estimation.
%
Performance on the in-the-wild test set is noticeably worse. This is likely due to the greater variability in camera types, viewpoints, and capture conditions in the in-the-wild data, which are not well represented in ZOD.

\textbf{OpenCVL (ZOD + Mapillary) as training data:}
Training on the full OpenCVL dataset improves both localization and orientation-estimation performance on the in-the-wild test set. 
Notably, incorporating this diverse and noisier training data maintains performance on the cross-area and snowy test sets derived from ZOD, and improves it when a downweighting factor is applied to the loss on Mapillary samples. 
This suggests that exposure to heterogeneous viewpoints and sensing conditions enhances overall robustness rather than causing overfitting to noisy supervision.

Still, the smaller gains observed without the downweighting factor underscore the need to account for uncertainty in the training data.
Nevertheless, the performance remains far from satisfactory, highlighting the difficulty of fine-grained cross-view localization in unconstrained real-world settings and the importance of evaluating on diverse datasets such as OpenCVL.

\textbf{Generalization from OpenCVL to KITTI:}
For completeness, we also evaluate the OpenCVL-trained model on the KITTI cross-area test split and compare it with a model trained on KITTI itself. All models are trained and evaluated without an orientation prior to ensure a fair comparison.

\begin{figure}[t]
    \centering
    \begin{minipage}{0.48\linewidth}
        \centering
        \includegraphics[width=\linewidth]{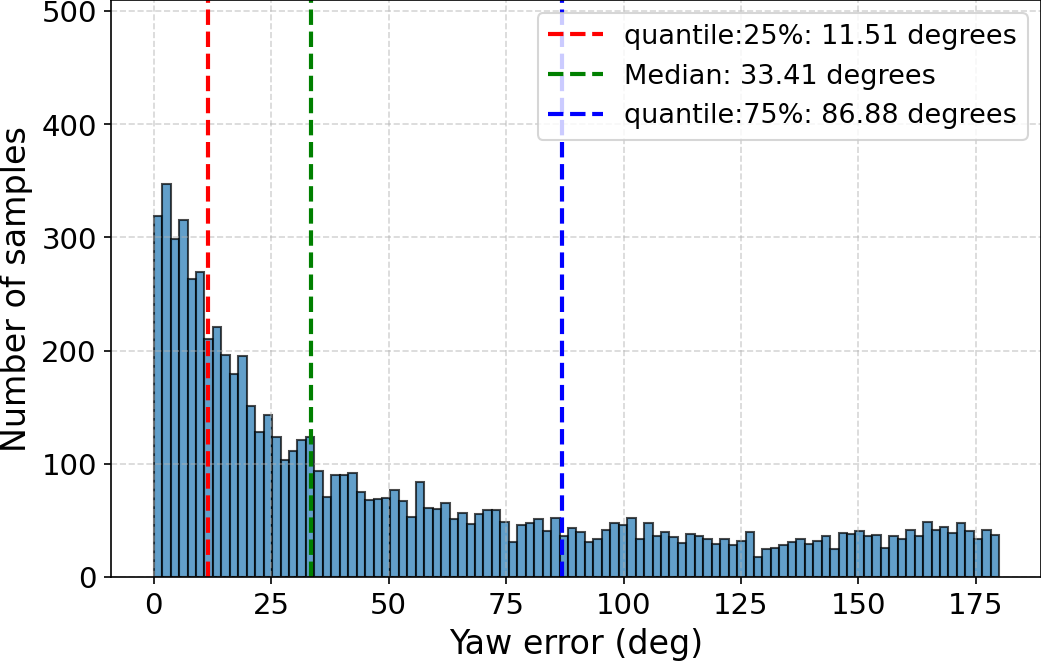}\\[-2pt]
        {\scriptsize Loc$^2$ trained on KITTI, tested on KITTI}
    \end{minipage}
    \hfill
    \begin{minipage}{0.48\linewidth}
        \centering
        \includegraphics[width=\linewidth]{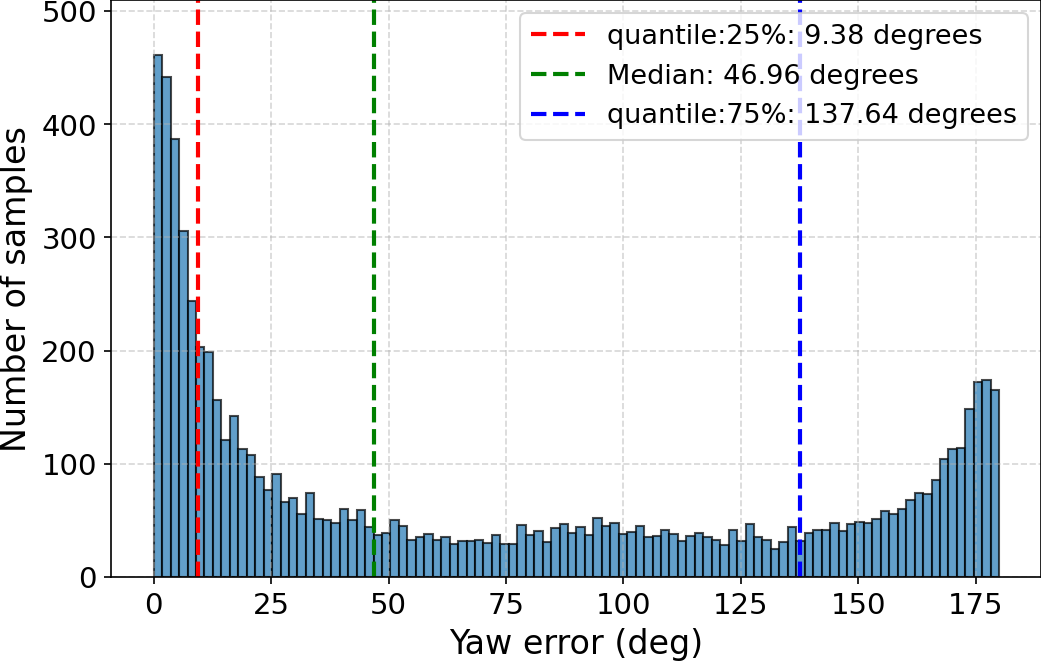}\\[-2pt]
        {\scriptsize Loc$^2$ trained on OpenCVL, tested on KITTI}
    \end{minipage}

    \caption{Orientation error histograms on the KITTI cross-area test split.}
    \label{fig:orientation_error_histogram}
\end{figure}

KITTI follows a different data distribution from OpenCVL. Nevertheless, the model trained on OpenCVL achieves localization accuracy comparable to that of the KITTI-trained model, as shown in Tab.~\ref{tab:kitti_cross_area_no_prior}. 
Compared with training on the ZOD subset alone, adding diverse Mapillary images again improves both localization and orientation prediction, despite the noise in their pose labels.
For orientation estimation, however, we observe that the OpenCVL-trained model produces more opposite-direction predictions, corresponding to errors of approximately $180^\circ$, see Fig.~\ref{fig:orientation_error_histogram}. 
Since KITTI images are mostly captured along roads, correctly distinguishing the driving direction is particularly important for minimizing orientation error. We hypothesize that models trained directly on KITTI learn stronger domain-specific cues for disambiguating forward and opposite driving directions. 
Even so, the OpenCVL-trained model produces a comparable number of mid-range errors and more low-error predictions.

Overall, these results show that OpenCVL supports reasonable cross-dataset generalization, but also confirm that robust cross-dataset transfer remains a challenging problem requiring further methodological advances.

\begin{table}[t]
\centering
\caption{KITTI cross-area evaluation without an orientation prior. For localization, the model trained on our full OpenCVL dataset generalizes slightly better to KITTI cross-area test set than a model trained on the KITTI distribution itself. The results ``KITTI (w/o ori. prior)'' are taken from the original Loc$^2$ paper~\cite{xia2025loc}. Best in bold.}
\label{tab:kitti_cross_area_no_prior}
\scriptsize 
\begin{tabular}{lcccc}
\toprule
Training data & Mean Loc. (m) & Med. Loc. (m) & Mean Ori. ($^\circ$) & Med. Ori. ($^\circ$) \\
\midrule
KITTI (w/o ori. prior) 
& 11.71 & \textbf{9.11} & \textbf{55.18} & \textbf{33.41} \\

OpenCVL (ZOD only) 
& 12.04 & 10.34 & 75.00 & 59.54 \\

OpenCVL 
& \textbf{11.25} & 9.40 & 71.17 & 46.96 \\
\bottomrule
\end{tabular}
\end{table}
\subsection{Qualitative Results}

\begin{figure}[t]
\centering

\begin{subfigure}{0.49\linewidth}
\qualimgsrc{\linewidth}{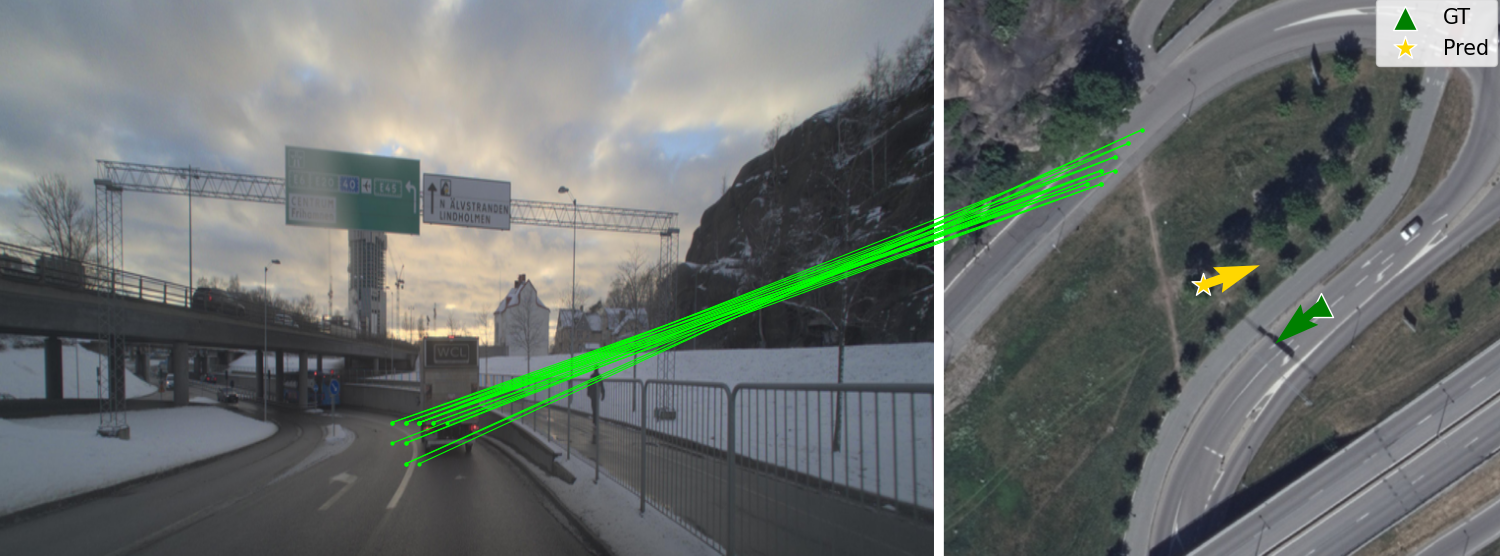}{Zenseact AB}{Lantm\"ateriet}
\caption{Snowy test, \textbf{trained on KITTI}}
\end{subfigure}
\hfill
\begin{subfigure}{0.49\linewidth}
\qualimgsrc{\linewidth}{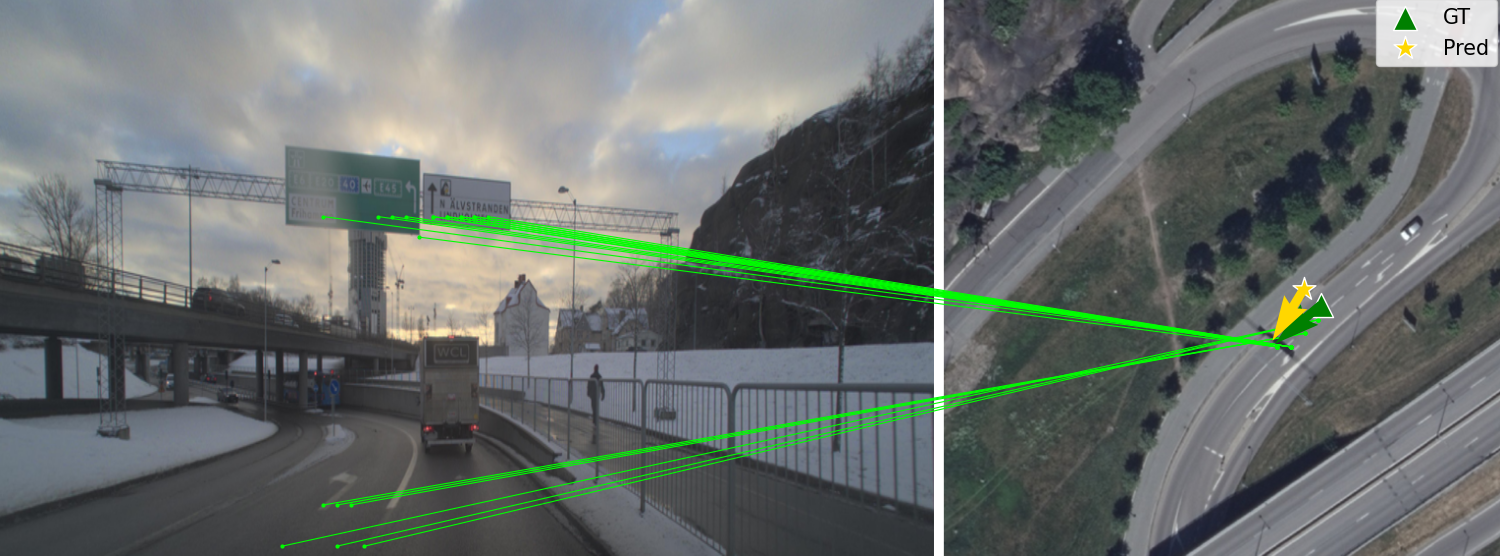}{Zenseact AB}{Lantm\"ateriet}
\caption{Snowy test, \textbf{trained on OpenCVL}}
\end{subfigure}
\smallskip

\begin{subfigure}{0.49\linewidth}
\qualimgsrc{\linewidth}{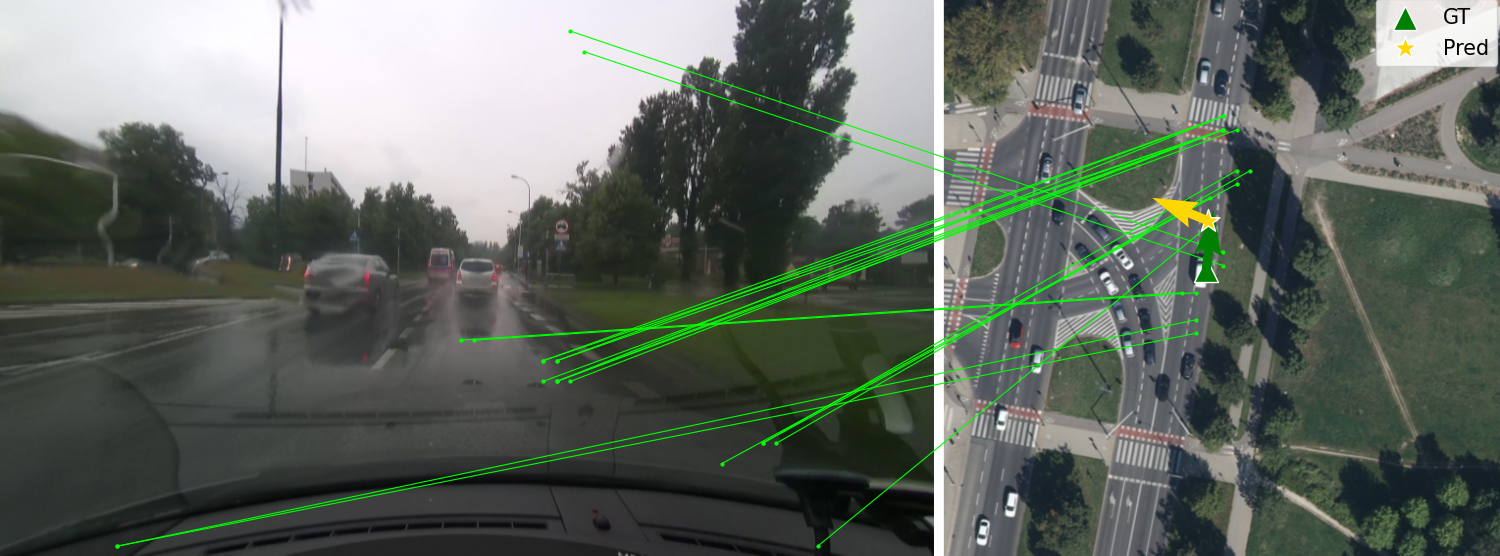}{Mapillary/grobin}{GUGiK}
\caption{In-the-wild test, \textbf{trained on KITTI}}
\end{subfigure}
\hfill
\begin{subfigure}{0.49\linewidth}
\qualimgsrc{\linewidth}{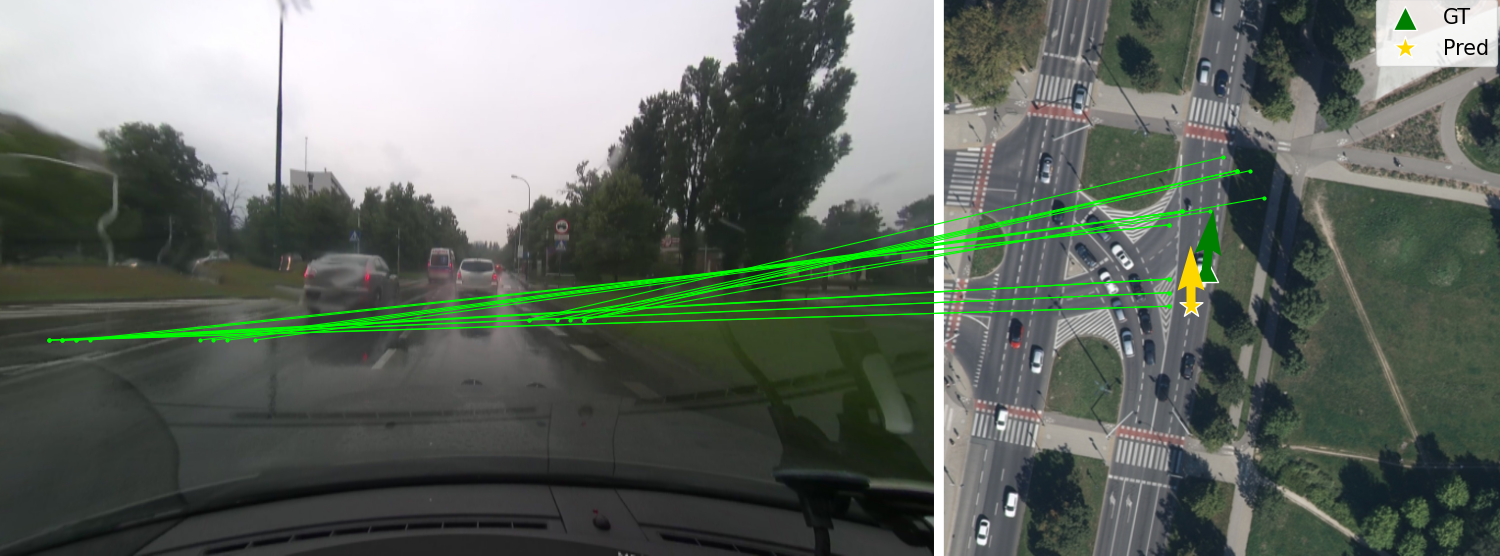}{Mapillary/grobin}{GUGiK}
\caption{In-the-wild test, \textbf{trained on OpenCVL}}
\end{subfigure}

\smallskip

\begin{subfigure}{0.49\linewidth}
\qualimgsrc{\linewidth}{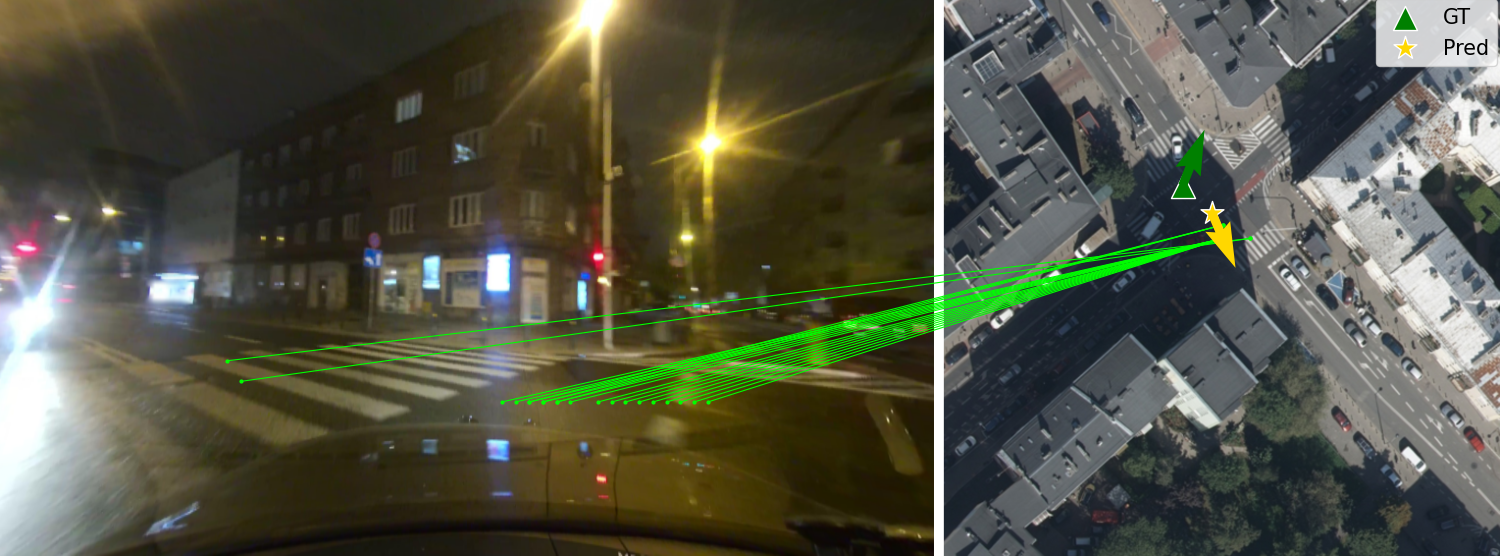}{Mapillary/grobin}{GUGiK}
\caption{fail: In-the-wild test, \textbf{trained on OpenCVL}}
\end{subfigure}
\hfill
\begin{subfigure}{0.49\linewidth}
\qualimgsrc{\linewidth}{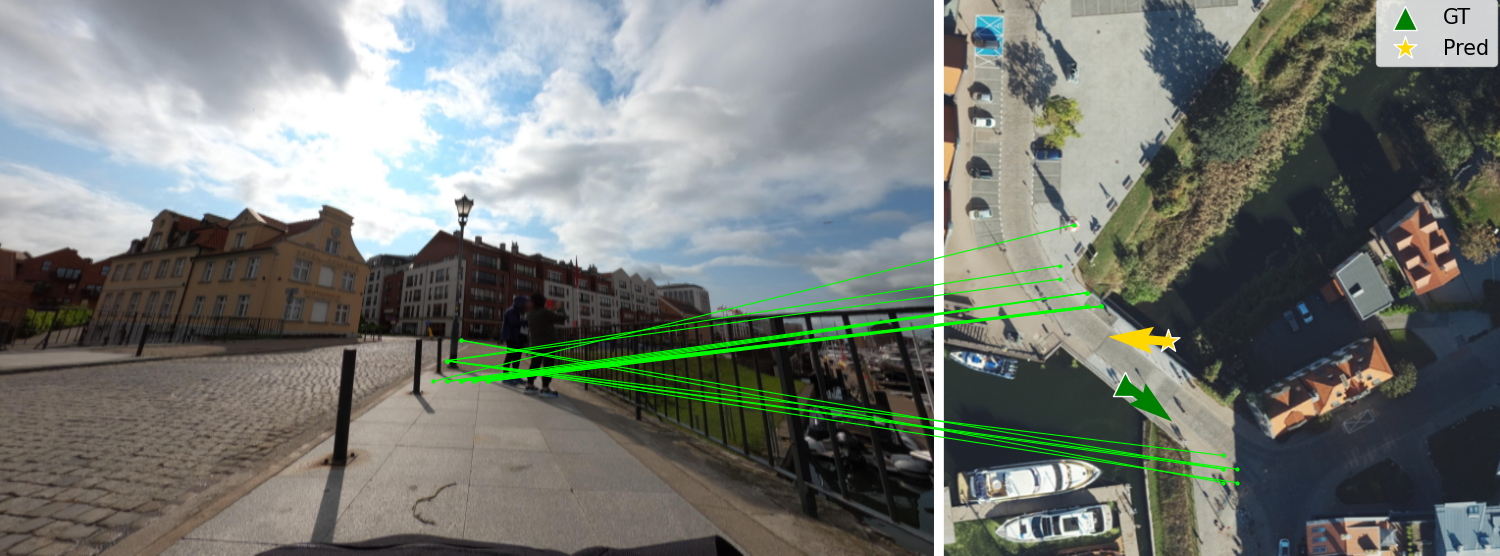}{Mapillary/masterofnoroad}{GUGiK}
\caption{fail: In-the-wild test, \textbf{trained on OpenCVL}}
\end{subfigure}

\caption{Qualitative results of Loc$^2$ across different test sets (green/yellow arrow: true/predicted pose). Lines show Loc$^2$'s top 20 correspondences by matching score.}
\label{fig:qualitative_results}
\end{figure}

We visualize qualitative results of the model trained on OpenCVL, as well as on KITTI.
Fig.~\ref{fig:qualitative_results} shows the localization and local feature matching results on the snowy and in-the-wild test sets.
The examples (a)+(b) and (c)+(d) compare training on KITTI to training on OpenCVL.
As shown in \ref{fig:qualitative_results}(a)+(b), training with KITTI often misses features on overhead signs,
which is corrected by training on OpenCVL.
On the in-the-wild examples of \ref{fig:qualitative_results}(c)+(d), the KITTI model struggles on aerial imagery with complex structures, compared to using OpenCVL.
Finally, the in-the-wild examples (e–f) illustrate still challenging scenarios in OpenCVL due to the great variability in camera orientation and location. In Fig.~\ref{fig:qualitative_results}(e), when the camera views a corner of an intersection, the model incorrectly localizes it to a different corner.
Fig.~\ref{fig:qualitative_results}(f) shows an image taken on a sidewalk. 
Although the model predicts a location on the sidewalk, it corresponds to an incorrect position and the opposite direction of travel.
These examples highlight the challenges posed by unconstrained viewpoints and diverse capture conditions, which are absent from prior benchmarks.

\subsection{Impact of Mapillary Pose Quality on CVL Training}
\label{sec:impactofmapillarydatacuration}
\begin{table}[t]
\centering
\caption{The impact of different pose labels for Mapillary images. Best in bold.}
\label{tab:mapillary_curation}
\scriptsize
\begin{tabular}{lcccc}
\toprule
Finetuning labels & Mean Loc. (m) & Med. Loc. (m) & Mean Ori. ($^\circ$) & Med. Ori. ($^\circ$) \\
\midrule
No finetuning & 8.05 & 5.75 & 26.73 & 10.57 \\
Raw GNSS pose & 7.41 & 5.73 & 22.09 & 7.70 \\
OpenSfM pose & 7.20 & 5.23 & 23.44 & 9.27 \\
\hline
\textbf{Our corrected pose} & \textbf{6.92} & \textbf{5.14} & \textbf{17.98} & \textbf{7.01} \\
\bottomrule
\end{tabular}
\end{table}

\begin{figure}[t]
    \centering
    \setlength{\tabcolsep}{2pt}

    \begin{tabular}{cccc}
        \imgsrc{0.23\linewidth}{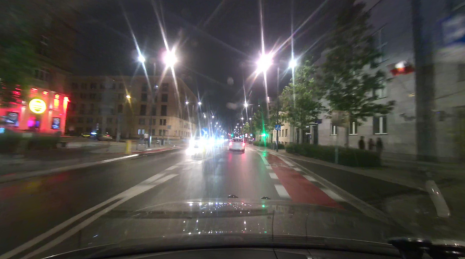}{Mapillary/grobin} &
        \imgsrc{0.23\linewidth}{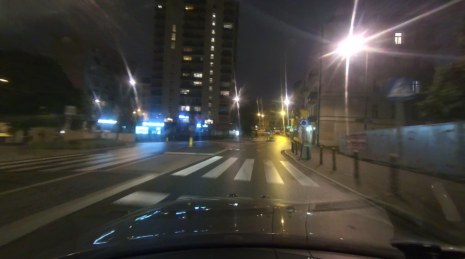}{Mapillary/grobin} &
        \imgsrc{0.23\linewidth}{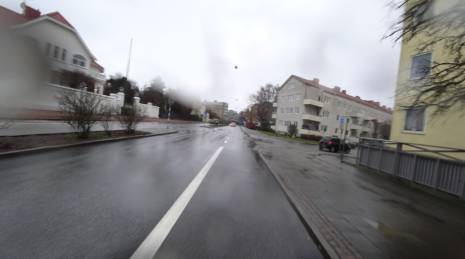}{Mapillary/movebybike} &
        \imgsrc{0.23\linewidth}{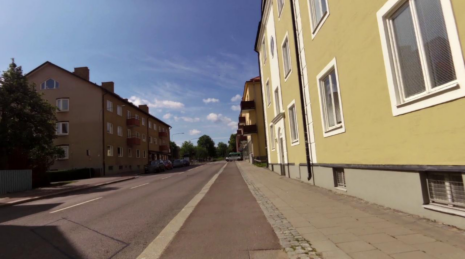}{Mapillary/adam\_sjoklin} \\[0.5pt]

        \imgsrc{0.23\linewidth}{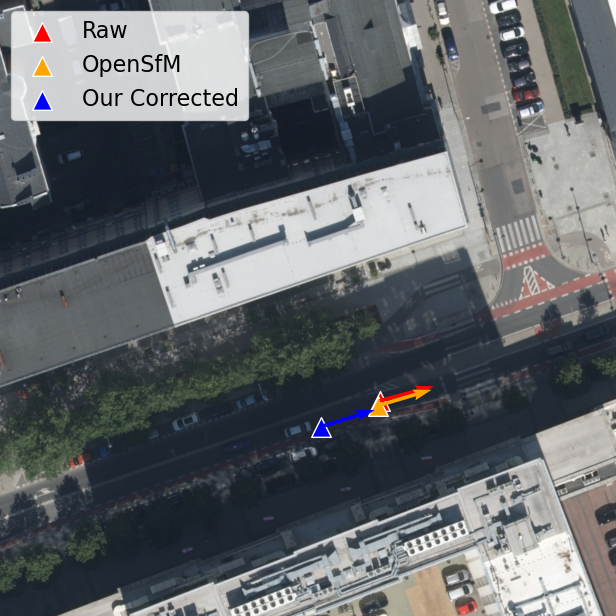}{GUGiK} &
        \imgsrc{0.23\linewidth}{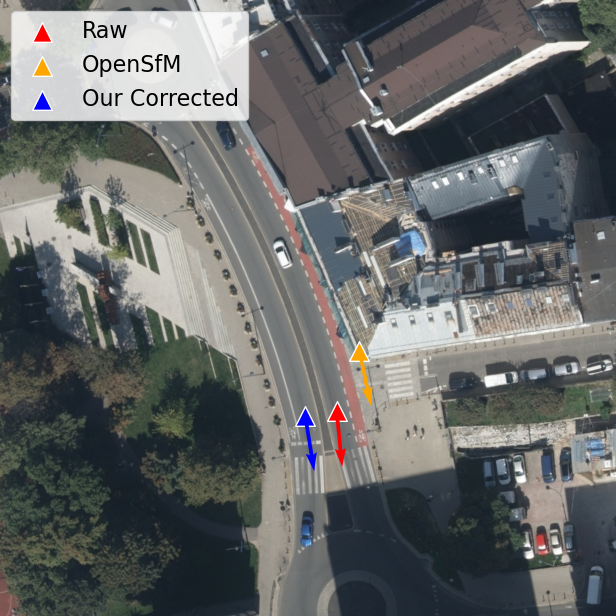}{GUGiK} &
        \imgsrc{0.23\linewidth}{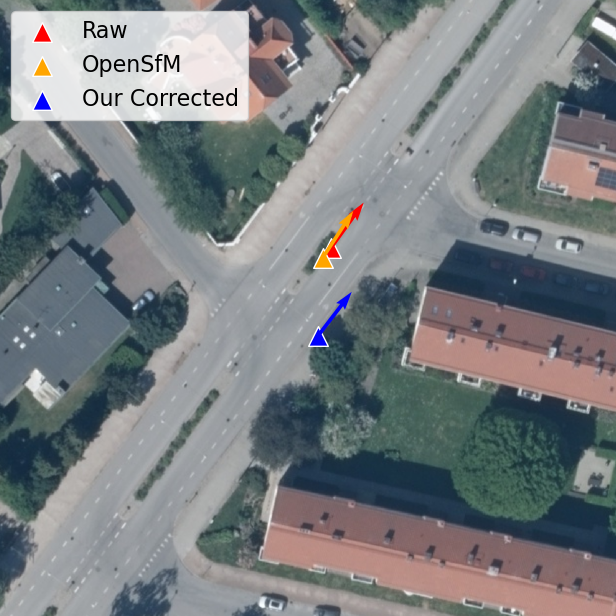}{Lantm\"ateriet} &
        \imgsrc{0.23\linewidth}{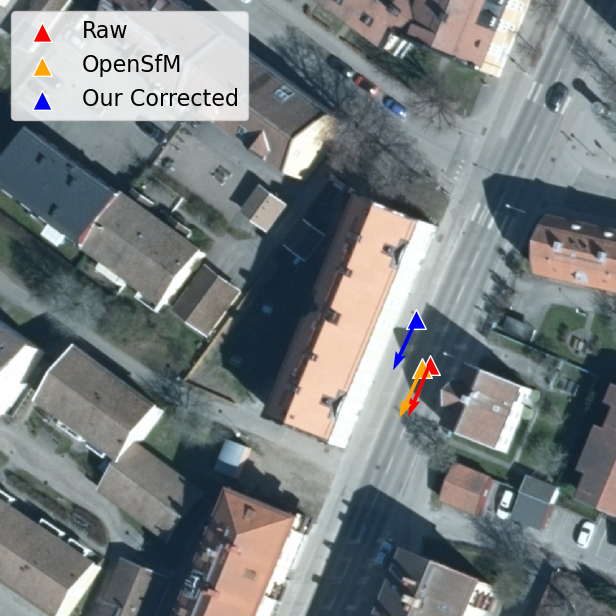}{Lantm\"ateriet}
    \end{tabular}

    \caption{Visual comparisons of the different pose annotations available for Mapillary images in our OpenCVL evaluation. The corrected Mapillary pose is significantly better than the raw Mapillary GNSS tag and the Mapillary-reported OpenSfM corrected pose.}
    \label{fig:mapillary_poses}
\end{figure}

Next, we report the benefits of our corrected Mapillary poses (before the final manual verification step in Sec.~\ref{sec:test_splits}) over the Mapillary raw GNSS tags and the Mapillary OpenSfM-refined poses.
For this experiment, we split the OpenCVL in-the-wild test set into two disjoint parts: one used for fine-tuning the ZOD-only pre-trained model with changing pose annotations, and the other used for validation. The validation poses are manually verified to ensure reliable evaluation.

As shown in Tab.~\ref{tab:mapillary_curation}, fine-tuning with Mapillary data always improves performance compared to the pre-trained model. 
However, the quality of the pose annotations determines the level of downstream benefits. While both raw GNSS tags and OpenSfM-refined poses provide some improvements, their noise leads to suboptimal supervision. In contrast, training with our corrected poses achieves the lowest localization and orientation errors. 
We show visual comparisons of the different pose annotations in Fig.~\ref{fig:mapillary_poses}.
The Mapillary poses (both raw and OpenSfM-refined) can contain obvious errors, such as placing the camera in the wrong lane or drifting from a bicycle lane to a car lane, and are therefore insufficient as reliable test data without our proposed correction framework.

\subsection{Leveraging OpenCVL for KITTI Training}
Finally, we investigate whether using OpenCVL can improve performance on the KITTI benchmark. 
We consider two ways of incorporating OpenCVL: joint training with KITTI and pre-training on OpenCVL followed by fine-tuning on KITTI. For KITTI samples, we follow the standard setting and use an orientation prior of $\pm10^\circ$, while the orientation remains unknown for OpenCVL samples.

As shown in Tab.~\ref{tab:kitti}, joint training on KITTI and OpenCVL slightly improves localization accuracy on the KITTI cross-area test sets, but increases orientation error. 
This is likely because KITTI samples are trained with a $\pm10^\circ$ orientation prior, whereas OpenCVL samples are trained without one. 
We therefore also evaluate OpenCVL pre-training followed by KITTI fine-tuning. This strategy yields the best overall performance, suggesting that OpenCVL provides useful large-scale and diverse pre-training data, while KITTI fine-tuning adapts the model to the target distribution and orientation prior.

\begin{table}[t]
\centering
\caption{KITTI evaluation with a $\pm10^\circ$ orientation prior. Best in bold.}
\label{tab:kitti}
\scriptsize
\begin{tabular}{lcccccccc}
\toprule
& \multicolumn{4}{c}{Same-area} 
& \multicolumn{4}{c}{Cross-area} \\
\cmidrule(lr){2-5} \cmidrule(lr){6-9}
Training strategy
& \multicolumn{2}{c}{$\downarrow$ Localization (m)} 
& \multicolumn{2}{c}{$\downarrow$ Orientation ($^\circ$)}
& \multicolumn{2}{c}{$\downarrow$ Localization (m)} 
& \multicolumn{2}{c}{$\downarrow$ Orientation ($^\circ$)} \\
\cmidrule(lr){2-3} \cmidrule(lr){4-5} 
\cmidrule(lr){6-7} \cmidrule(lr){8-9}
& Mean & Med. & Mean & Med. 
& Mean & Med. & Mean & Med. \\
\midrule
KITTI only
& 1.13 & 0.77 & 1.97 & 1.43
& 5.60 & 3.01 & 3.32 & \textbf{2.12} \\

OpenCVL + KITTI
& 1.12 & 0.77 & 1.90 & 1.37
& 5.43 & 2.99 & 4.06 & 2.38 \\

OpenCVL $\rightarrow$ KITTI 
& \textbf{0.93} & \textbf{0.61} 
& \textbf{1.62} & \textbf{1.19}
& \textbf{5.07} & \textbf{2.84} 
& \textbf{3.30} & 2.15 \\
\bottomrule
\end{tabular}
\end{table}
\section{Conclusion}
\label{sec:conclusion}

We introduced OpenCVL, a large-scale, diverse, and open dataset for fine-grained cross-view localization. OpenCVL combines high-precision ground-truth data with in-the-wild imagery, providing diversity in collection platforms, temporal conditions, and viewpoints that is largely absent in prior datasets. 
It offers carefully curated evaluation splits, including cross-area, snowy, and in-the-wild test sets, enabling more realistic and comprehensive benchmarking of CVL methods. A dedicated framework was developed to create accurate labels for this in-the-wild imagery.
Importantly, all data in OpenCVL originates from permissive sources, ensuring that the dataset can be freely shared and extended. 
Our experiments  on the state-of-the-art Loc$^2$
show that incorporating noisy but diverse data can significantly improve localization performance, while the challenging in-the-wild test set reveals further opportunities for improvement and future research. We hope OpenCVL will facilitate future research toward scalable and robust cross-view localization systems.

\newpage
\section*{Appendix}

Here we provide supplementary material to support the main paper:

\begin{enumerate}[label=\Alph*.]
    \item The complete list of city–country pairs included in OpenCVL.
    \item Extra baseline methods on OpenCVL.
    \item Discussion on Orthophoto vs. True Orthographic Imagery.
    \item More qualitative visualizations of our pose correction pipeline.
\end{enumerate}

\subsection*{A. Complete City--Country List of OpenCVL}
As discussed in Sec.~3 of the main paper, OpenCVL contains ground-level images sourced from the Zenseact Open Dataset (ZOD)~\cite{alibeigi2023zenseact} and Mapillary~\cite{Mapillary}. 
The Mapillary subset consists of 342,901 images in total, including a test set of 1,361 images that are geographically close to the ZOD data, and a training set of 341,540 images. 
The training images cover 41 cities across four European countries: Sweden, Poland, Norway, and the Netherlands. 
We provide the complete list of city--country pairs included in the training set in Tab.~\ref{tab:city_samples}.
For each city, we define a rectangular region within the urban area, divide it into uniform tiles of approximately 1~km$^2$, and download a fixed set of 100 images per tile whenever enough images are available.

\begin{table}[t]
\centering
\scriptsize
\begin{tabular}{llr}
\toprule
Country & City & Samples \\
\midrule
NL & Amsterdam & 21664 \\
NL & Breda & 4848 \\
NL & Delft & 2174 \\
NL & Eindhoven & 6774 \\
NL & Groningen & 5626 \\
NL & Haarlem & 2750 \\
NL & Leiden & 1681 \\
NL & Maastricht & 5327 \\
NL & Nijmegen & 7356 \\
NL & Rotterdam & 21716 \\
NL & The Hague & 6316 \\
NL & Utrecht & 4516 \\
\midrule
NO & Alesund & 432 \\
NO & Bergen & 13019 \\
NO & Drammen & 1390 \\
NO & Fredrikstad & 1067 \\
NO & Kristiansand & 1377 \\
NO & Oslo & 13626 \\
NO & Stavanger & 4039 \\
NO & Tromso & 3848 \\
NO & Trondheim & 9305 \\
\midrule
PL & Bydgoszcz & 6008 \\
PL & Gdansk & 10538 \\
PL & Katowice & 8579 \\
PL & Krakow & 20227 \\
PL & Lodz & 10117 \\
PL & Lublin & 9077 \\
PL & Poznan & 25957 \\
PL & Szczecin & 6634 \\
PL & Wroclaw & 19954 \\
\midrule
SE & Goteborg & 23321 \\
SE & Handen & 986 \\
SE & Helsingborg & 1836 \\
SE & Linkoping & 4941 \\
SE & Lund & 1429 \\
SE & Malmo & 10988 \\
SE & Norrkoping & 946 \\
SE & Orebro & 4608 \\
SE & Stockholm & 20929 \\
SE & Uppsala & 10049 \\
SE & Vasteras & 5565 \\
\bottomrule
\end{tabular}
\caption{Number of samples per city in the dataset.}
\label{tab:city_samples}
\end{table}

\subsection*{B. Extra Baseline Methods on OpenCVL}
Apart from Loc$^2$~\cite{xia2025loc}, we also trained and tested a homography-based method, HC-Net, and a global descriptor-based method, CCVPE, on OpenCVL with the same data weighting. 

\begin{figure}[h]
    \centering
    \begin{minipage}{0.69\linewidth}
        \centering
        \imgsrc{\linewidth}{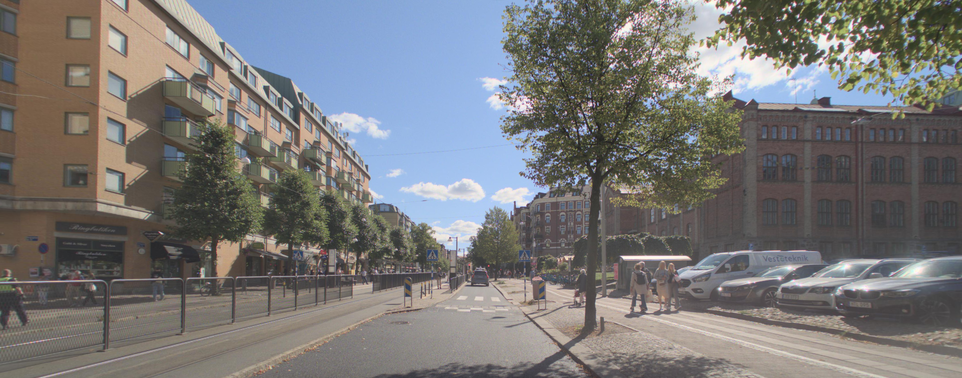}{Zenseact AB}\\[-2pt]
        {\scriptsize Original ground-level image}
    \end{minipage}
    \hfill
    \begin{minipage}{0.29\linewidth}
        \centering
        \includegraphics[width=\linewidth]{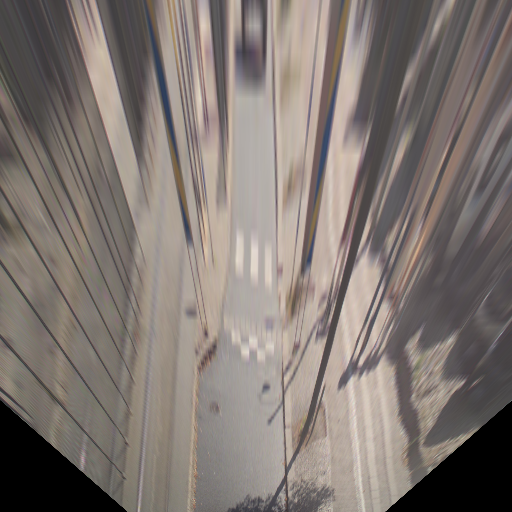}\\[-2pt]
        {\scriptsize BEV-warped image}
    \end{minipage}
    \caption{Example of the homography-based preprocessing used for HC-Net.}
    \label{fig:homography}
\end{figure}

\textbf{Settings:} For HC-Net, we first warp each ground-level image into a bird's-eye-view (BEV) perspective before feeding it to the model. We use the image-specific HFoV for the transformation, while the remaining homography parameters are selected by visually inspecting a small set of examples and then kept fixed for all samples, as illustrated in Fig.~\ref{fig:homography}.

CCVPE matches each limited-HFoV ground image to a sector of the aerial descriptor. Although OpenCVL ground images are captured by different cameras with potentially varying HFoVs, we assume a fixed HFoV of $120^\circ$ for all ground images when training CCVPE, and make minor architectural modifications to support our input resolution.

\textbf{Results:} Loc$^2$ consistently outperforms CCVPE and HC-Net in localization across all evaluation splits, confirming the advantage of local correspondence-based pose estimation for fine-grained CVL. 
For Loc$^2$ and CCVPE, the orientation error is lower on the Snowy set than on the other test sets, likely because Snowy images are mostly captured along roads and therefore provide stronger directional cues. 
In contrast, HC-Net performs poorly overall, especially for orientation estimation, indicating that homography-based matching is less robust for limited-HFoV ground images when the orientation is unknown. 

\begin{table}[h]
\centering
\caption{Extra baseline results. Best in bold.}
\scriptsize
\setlength{\tabcolsep}{2.0pt}
\renewcommand{\arraystretch}{0.85}
\resizebox{\columnwidth}{!}{%
\begin{tabular}{lcccccccccccc}
\toprule
& \multicolumn{4}{c}{Cross-area test set} 
& \multicolumn{4}{c}{Snowy test set} 
& \multicolumn{4}{c}{In-the-wild test set} \\

\cmidrule(lr){2-5}
\cmidrule(lr){6-9}
\cmidrule(lr){10-13}

& \multicolumn{2}{c}{Loc.} & \multicolumn{2}{c}{Ori.}
& \multicolumn{2}{c}{Loc.} & \multicolumn{2}{c}{Ori.}
& \multicolumn{2}{c}{Loc.} & \multicolumn{2}{c}{Ori.} \\

\cmidrule(lr){2-3}
\cmidrule(lr){4-5}
\cmidrule(lr){6-7}
\cmidrule(lr){8-9}
\cmidrule(lr){10-11}
\cmidrule(lr){12-13}

& Mean & Med. & Mean & Med.
& Mean & Med. & Mean & Med.
& Mean & Med. & Mean & Med. \\
\midrule
HC-Net
& 13.87 & 13.96 & 99.73 & 106.08
& 12.82 & 12.85 & 94.89 & 107.73
& 12.66 & 12.28 & 85.89 & 88.27 \\

CCVPE
& 11.95 & 7.64 & 25.59 & \textbf{4.56}
& 10.81 & 7.66 & 7.41 & \textbf{2.48}
& 16.24 & 12.51 & 38.64 & \textbf{7.88} \\

Loc$^2$
& \textbf{6.72} & \textbf{4.30} & \textbf{11.52} & 5.24
& \textbf{7.80} & \textbf{5.48} & \textbf{5.09} & 3.52
& \textbf{7.90} & \textbf{6.19} & \textbf{26.07} & 12.01 \\
\bottomrule
\end{tabular}%
}
\end{table}

\subsection*{C. Discussion on Orthophoto vs. True Orthographic Imagery}
In this work, we use the term aerial imagery to refer to the overhead images employed for cross-view localization. In practice, most publicly available overhead imagery is distributed as orthophotos, that is, orthorectified aerial images. These images are generated from aerial photographs that have been geometrically corrected to remove perspective distortions caused by camera tilt, terrain relief, and sensor geometry, resulting in imagery with a consistent true metric scale.

True orthographic imagery represents an ideal nadir-view projection in which all scene points are observed from a perfectly vertical viewpoint, without perspective distortions or occlusions. In contrast, orthophotos are generated by projecting aerial photographs onto a digital elevation model (DEM) and resampling them onto a map grid. While this procedure substantially reduces geometric distortions, the resulting images are not strictly orthographic, and building facades may still be visible, especially in dense urban areas.

Note that the Polish national mapping agency~\cite{Poland} provides true orthographic imagery for limited areas. 
Our dataset includes such imagery for the city of Poznan, meaning that for this city we have both standard aerial imagery (orthophotos) and true orthographic images. 
This enables future research to study the potential impact of these different overhead representations. 
In our experiments, however, we use orthophotos for consistency across all locations.

\subsection*{D. More Qualitative Examples of Our Pose Correction Pipeline}
\begin{figure}
    \centering
    \includegraphics[width=0.99\linewidth]{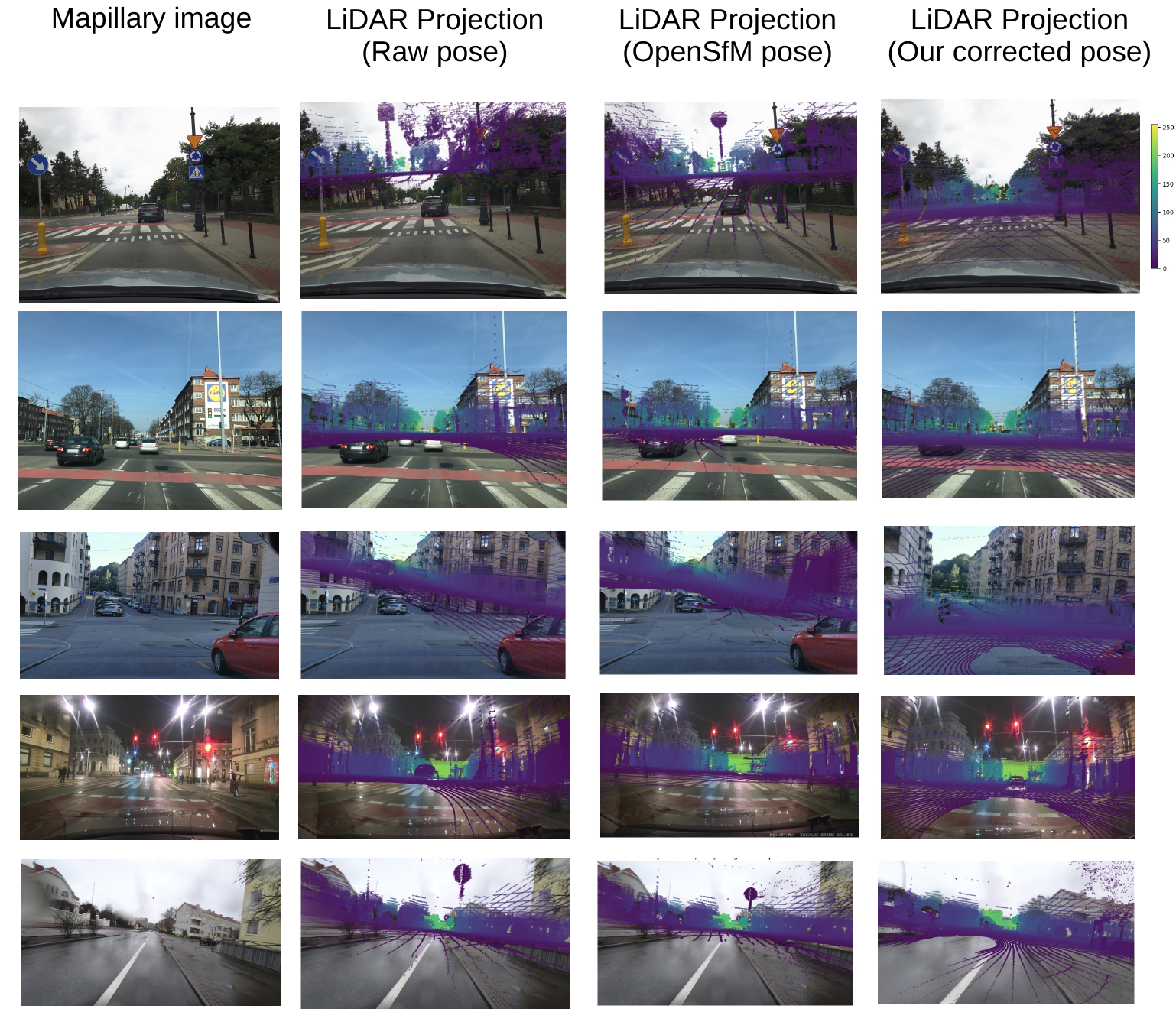}
    \caption{Visualization of ZOD LiDAR pointcloud projected into the Mapillary images using the raw Mapillary pose, the OpenSfM pose, and our corrected pose. The structure in LiDAR pointcloud overlaps well with the structure in the images when projected using our corrected pose for the Mapillary images.}
    \label{fig:visualization_mapi_correction}
\end{figure}

We show in Fig.~\ref{fig:visualization_mapi_correction} more visualizations of the LiDAR pointcloud projection from ZOD into nearby Mapillary images. The projection is performed using either the raw pose from Mapillary, the OpenSfM pose from Mapillary, or the corrected pose from our pipeline. The camera intrinsics and distortion parameters are used as originally reported by Mapillary for the former two, and in the case of latter are computed by our pipeline. If the Mapillary pose would be accurate, the structure in LiDAR pointcloud (poles, traffic signs, buildings, etc.) would overlap well with the structure visible in the image. These visualizations therefore show the quality of our pose estimation pipeline, and thereby the quality of ground-truth poses in our challenging OpenCVL in-the-wild test set.

\section*{Acknowledgements}
This work was supported by the Swiss Open Research Data Fund (CHORD), project number PgB\_25-28\_674\_A1\_19. We thank Yejie Guo and Bakul Jangley for their contributions to the initial phase of this project.

%
%
\bibliographystyle{splncs04}
\bibliography{main}
\end{document}